\documentclass{article} % For LaTeX2e
\usepackage{iclr2025_conference,times}

\usepackage{amsmath,amsfonts,bm}

\def\figref#1{figure~\ref{#1}}
\def\secref#1{section~\ref{#1}}
\def\eqref#1{equation~\ref{#1}}
\def\1{\bm{1}}

\DeclareMathAlphabet{\mathsfit}{\encodingdefault}{\sfdefault}{m}{sl}
\SetMathAlphabet{\mathsfit}{bold}{\encodingdefault}{\sfdefault}{bx}{n}

\usepackage{hyperref}
\usepackage{url}

\usepackage[utf8]{inputenc} % allow utf-8 input
\usepackage[T1]{fontenc}    % use 8-bit T1 fonts
\usepackage{hyperref}       % hyperlinks
\usepackage{url}            % simple URL typesetting
\usepackage{booktabs}       % professional-quality tables
\usepackage{amsfonts}       % blackboard math symbols
\usepackage{nicefrac}       % compact symbols for 1/2, etc.
\usepackage{microtype}      % microtypography
\usepackage{xcolor}         % colors
\usepackage{multirow}      % multirow for table

\usepackage{algorithm}  
\usepackage{algorithmicx}  
\usepackage{algpseudocode}  
\usepackage{amsmath}
\usepackage{pifont}
\usepackage{enumitem}
\usepackage{graphicx}
\usepackage{subfigure}
\usepackage{bm}
\usepackage{comment}
\usepackage{CJKutf8}
\usepackage{bbding}
\usepackage{diagbox}
\usepackage{wrapfig}
\usepackage{duckuments}
\usepackage{stfloats}

\definecolor{MyDarkBlue}{rgb}{0,0.5,1}
\definecolor{MyDarkGreen}{rgb}{0.02,0.6,0.02}
\definecolor{MyDarkRed}{rgb}{0.8,0.02,0.02}
\definecolor{MyDarkOrange}{rgb}{0.40,0.2,0.02}
\definecolor{MyYellow}{rgb}{1,0.55,0}
\definecolor{MyPurple}{RGB}{111,0,255}
\definecolor{MyRed}{rgb}{1.0,0.0,0.0}
\definecolor{MyGold}{rgb}{0.75,0.6,0.12}
\definecolor{MyDarkgray}{rgb}{0.66, 0.66, 0.66}
\definecolor{default}{RGB}{0,0,0}

\newcommand\eg{\textit{e.g., }}
\newcommand\ie{\textit{i.e., }}

\newcommand{\eq}[1]{\begin{equation}#1\end{equation}}
\newcommand{\model}{LS-Imagine} % model name
\renewcommand{\eqref}[1]{Eq.~(\ref{#1})} %公式引用
\renewcommand{\figref}[1]{Figure~\ref{#1}} % 图片引用
\newcommand{\tabref}[1]{Table~\ref{#1}} %表格引用
\renewcommand{\secref}[1]{Sec.~\ref{#1}} %章节引用
\newcommand{\appref}[1]{\underline{Appendix~\ref{#1}}} %附录引用
\newcommand{\reb}[1]{{\color{black}{#1}}}
\newcommand{\mycheckmark}{{\textcolor{MyDarkGreen}{\checkmark}}}
\newcommand{\myxmark}{{\textcolor{MyDarkRed}{\ding{55}}}}

\title{Open-World Reinforcement Learning over\\Long Short-Term Imagination}

\author{
Jiajian Li$^{1}$\thanks{Equal contribution.} 
\quad
Qi Wang$^{1,2*}$ 
\quad
Yunbo Wang$^{1}$\thanks{Corresponding author: Yunbo~Wang \texttt{<yunbow@sjtu.edu.cn>}.}
\quad
Xin Jin$^{2}$
\quad
Yang Li$^{3}$
\quad
Wenjun Zeng$^{2}$
\quad
Xiaokang Yang$^{1}$\\
$^1$ MoE Key Lab of Artificial Intelligence, AI Institute, Shanghai Jiao Tong University, Shanghai, China\\
$^2$ Ningbo Institute of Digital Twin, Eastern Institute of Technology, Ningbo, China\\
$^3$ School of Computer Science and Technology, East China Normal University, Shanghai, China\\
\textcolor{magenta}{\url{https://qiwang067.github.io/ls-imagine}}
}

\iclrfinalcopy % Uncomment for camera-ready version, but NOT for submission.

\begin{document}

\maketitle

\begin{abstract}

Training visual reinforcement learning agents in a high-dimensional open world presents significant challenges. While various model-based methods have improved sample efficiency by learning interactive world models, these agents tend to be ``short-sighted'', as they are typically trained on short snippets of imagined experiences. We argue that the primary challenge in open-world decision-making is improving the exploration efficiency across a vast state space, especially for tasks that demand consideration of long-horizon payoffs. In this paper, we present \model{}, which extends the imagination horizon within a limited number of state transition steps, enabling the agent to explore behaviors that potentially lead to promising long-term feedback. The foundation of our approach is to build a \textit{long short-term world model}. To achieve this, we simulate goal-conditioned jumpy state transitions and compute corresponding affordance maps by zooming in on specific areas within single images. This facilitates the integration of direct long-term values into behavior learning. Our method demonstrates significant improvements over state-of-the-art techniques in MineDojo.

% Training model-based reinforcement learning~(MBRL) models with raw pixels in the open-world environment face two significant challenges, \ie the struggle with complex dynamics and the low sample efficiency in vast state space.
% %
% To cope with these challenges, we present a novel long short-term imagination approach named \model{}. The key insight is to leverage \textit{long-term transition} in the latent imagination of the world model with intrinsic rewards, while simultaneously relying on \textit{short-term} latent imagination. 
% %
% Concretely, to employ long-term transition, we learn and adopt the affordance map to indicate the relevance of task instruction defined by text and particular areas of visual observation. 
% %
% In doing so, we encourage the world model to learn behavior on long-term transition branches, and the agent can explore task-conditioned behaviors on the long-term latent imagination of the world model.
% %
% In 5 tasks from the MineDojo benchmark, \model{} outperforms other baselines, demonstrating its effectiveness in challenging open world. 

% improves the Dream-to-Control framework~\citep{hafner2023dreamerv3} in two aspects.  
%别提 mineclip
% First, we leverage MineCLIP to generate affordance maps for producing task-oriented intrinsic rewards,.
%可以简单带一下 dense reward
%
% Second, for each state, the jumpy imagination predictor is utilized to decide \textit{long-term} or \textit{short-term imagination} in representation learning of the world model.
%我们做的不错

\end{abstract}

\vspace{-3pt}
\section{Introduction}
\vspace{-3pt}
Open-world decision-making in the context of reinforcement learning (RL) involves the following characteristics: 
(i) The agent operates within an interactive environment that features a vast state space; 
(ii) The learned policy presents a high degree of flexibility, allowing interaction with various objects in the environment; 
(iii) The agent lacks full visibility of the internal states and physical dynamics of the external world, meaning that its perception of the environment (\textit{e.g.}, raw images) carries substantial uncertainty. 
For example, Minecraft serves as a typical open-world game.

Building upon recent progress in visual control, open-world decision-making aims to train agents to approach human-level intelligence based solely on high-dimensional visual observations. However, this presents significant challenges.
For example, in Minecraft tasks, existing methods like Voyager~\citep{wang2024voyager} employ specific Minecraft APIs as the high-level controller, which is incompatible with standard visual control settings. 
While approaches such as PPO-with-MineCLIP~\citep{fan2022minedojo} and DECKARD~\citep{nottingham2023embodied} perform low-level visual control, these model-free RL methods struggle to grasp the underlying mechanics of the environment. This may result in high trial-and-error costs, leading to inefficiencies in both exploration and sample usage. 
Although DreamerV3~\citep{hafner2023dreamerv3} employs a model-based RL (MBRL) approach to improve sample efficiency, it is often ``short-sighted'' since the policy is optimized using short-term experiences---typically $15$ time steps---generated by the world model. The absence of long-term guidance significantly hampers an effective exploration of the vast solution space of the open world.

\begin{figure}[t]
\vspace{-5pt}
    \centering
    \includegraphics[width=\textwidth]{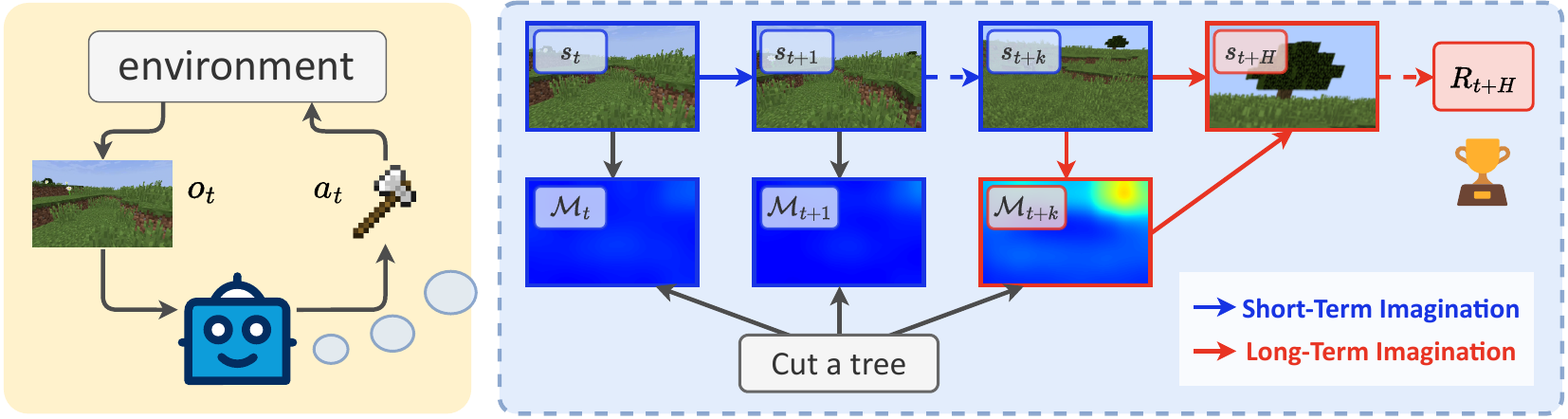}
    \vspace{-18pt}
    \caption{
    The general framework of \model{}, an MBRL agent that operates solely on raw pixels. The fundamental idea is to extend the imagination horizon within a limited number of state transition steps, enabling the agent to explore behaviors that potentially lead to promising long-term feedback.
     }
    \label{fig:ls_imagine_framework}
    \vspace{-10pt}
\end{figure}

To improve the behavior learning efficiency of MBRL, in this paper, we introduce a novel method named Long Short-Term Imagination (\model{}).
Our key approach involves \emph{enabling the world model to efficiently simulate the long-term effects of specific behaviors without the need for repeatedly rolling out one-step predictions}. 
As illustrated in \figref{fig:ls_imagine_framework}, once trained, the world model provides both instant and jumpy state transitions\footnote{As shown in \figref{fig:ls_imagine_framework}, a jumpy transition allows the agent to bypass intermediate states and directly simulate a task-relevant future state $s_{t+H}$ in one step. This process occurs exclusively during world model imagination.} along with corresponding (intrinsic) rewards, facilitating policy optimization in a joint space of short- and long-term imaginations. This encourages the agent to explore behaviors that lead to promising long-term outcomes.

The foundation of \model{} is to train a \textit{long short-term world model}, which requires integrating task-specific guidance into the representation learning phase based on off-policy experience replay.
However, this creates a classic ``chicken-and-egg'' dilemma: \emph{without true data showing the agent has reached the goal, how can we effectively train the model to simulate jumpy transitions from current states to pivotal future states that suggest a high likelihood of achieving that goal?}
To address this issue, we first continuously zoom in on individual images to simulate the consecutive video frames as the agent approaches the goal. 
We then generate affordance maps\footnote{
Affordance maps highlight regions within an observation that are potentially relevant to the task~\citep{qi2020learning,wang2022adaafford}.} by evaluating the relevance of the pseudo video to task-specific goals presented in textual instructions, using the established MineCLIP reward model~\citep{fan2022minedojo}.
Subsequently, we train specific branches of the world model to capture both instant and jumpy state transitions, using pairs of image observations from adjacent time steps as well as those across longer intervals.
Finally, we optimize the agent's policy based on a finite sequence of imagined latent states generated by the world model, integrating a more direct estimate of long-term values into decision-making.

Let's use the example in \figref{fig:ls_imagine_framework} to further elaborate the novel aspects of the behavior learning process: 
After receiving the instruction ``cut a tree'', the agent simulates near-future states based on the current real observation. 
It initially performs several single-step rollouts until it identifies a point in time for a long-distance state jump that allows it to approach the tree. The agent then executes this jump and optimizes its policy network to maximize the long-sight value function.

We evaluate our approach in the challenging open-world tasks from MineDojo~\citep{fan2022minedojo}. 
\model{} demonstrates superior performance compared to existing visual RL methods.

The contributions of this work are summarized as follows:
\begin{itemize}[leftmargin=*]
\vspace{-5pt}\item  
We present a novel model-based RL method that captures both instant and jumpy state transitions and leverages them in behavior learning to improve exploration efficiency in the open world.
\vspace{-2pt}
\item
Our approach presents four concrete contributions: (i) a long short-term world model architecture, (ii) a method for generating affordance maps through image zoom-in, (iii) a novel form of intrinsic rewards based on the affordance map, and (iv) an improved behavior learning method that integrates long-term values and operates on a mixed long short-term imagination pathway.
\vspace{-5pt}
\end{itemize}

% \yb{Para 3: Give more detailed descriptions for model design.}
% \yb{Para 4: Summarize the key contributions and the experimental results}

\vspace{-3pt}\section{Problem Formulation and Notations}
\vspace{-3pt}

We solve visual reinforcement learning as a partially observable Markov decision process (POMDP), using MineDojo as the test bench. 
Specifically, our method manipulates low-dimensional control signals $a_t$ while receiving only sequential high-dimensional visual observations $o_{<t}$ and episodic sparse rewards $r^\text{env}$, without access to the internal APIs of the open-world games.
In comparison, as shown in \tabref{tab:setting_cmp}, existing Minecraft agents present notable distinctions in learning paradigms (\textit{i.e.}, controller), observation data, and the use of expert demonstrations.

The world model presented in this paper consists of two main components: a short-term transition branch and a long-term imagination branch. As a result, it employs a complex notation system. We now introduce the key notations that will be frequently used throughout the paper: 
\begin{itemize}[leftmargin=*]
\vspace{-5pt}
    \item $\mathcal{M}_t$ represents the affordance map.
    \vspace{-2pt}
    \item $c_t$ denotes the episode continuation flag.
    \vspace{-2pt}
    % \item $\mathcal{M}_t$ denotes the affordance map.
    % \vspace{-2pt}
    \item $j_t$ is the jumping flag that triggers jumpy state imaginations.
    \vspace{-2pt}
    \item $\Delta_t$ represents the number of environmental steps between the jumpy transitions.
    \vspace{-2pt}
    \item $G_t$ is the cumulative reward over $\Delta_t$.
    \vspace{-5pt}
\end{itemize}
We use $(o_t^{\prime},a_t^{\prime}, \mathcal{M}^{\prime}_t, {r}_t^{\prime}, c_t^{\prime},j_t^{\prime},\Delta_t^{\prime}, G_t^{\prime})$ to represent the simulated environment data that are used to train the long-term imagination branch of the world model.
The policy is learned on trajectories of mixed long- and short-term imaginations $\{(\hat{s}_t, \hat{a}_t, \hat{r}_t, \hat{c}_t, \hat{j}_t, \hat{\Delta}_t, \hat{G}_t) \}$, where $\hat{s}_t$ represents the latent state, and the variables predicted by the model are indicated using the superscript $(\ \hat{} \ )$.

\begin{table*}[t]
\vspace{-10pt}
\caption{Experimental setups of the Minecraft AI agents. \textit{IL} is short for imitation learning.} 
\label{tab:setting_cmp}
\vspace{-3pt}
\setlength\tabcolsep{4.2pt}
\begin{center}
% \begin{small}
\small
% \scriptsize
\centering
\begin{tabular}{lccc}
\toprule
Model & Controller & Observation & Video Demos  \\
\midrule
DECKARD~\citeyearpar{nottingham2023embodied}&RL &Pixels \& Inventory & \mycheckmark \\
Auto MC-Reward~\citeyearpar{li2024auto}&IL + RL &Pixels \& GPS & \myxmark \\
Voyager~\citeyearpar{wang2024voyager}& GPT-4 & Minecraft simulation \& Error trace& \myxmark \\ 
DEPS~\citeyearpar{wang2023describe}& IL &Pixels \& Yaw/pitch angle \& GPS \& Voxel & \myxmark \\ 
STEVE-1~\citeyearpar{lifshitz2023steve}& Generative model & Pixels &\myxmark\\ 
VPT~\citeyearpar{baker2022video}& IL + RL & Pixels & \mycheckmark \\
DreamerV3~\citeyearpar{hafner2023dreamerv3} &RL& Pixels & \myxmark \\
\model{} & RL &Pixels  & \myxmark \\
\bottomrule
\end{tabular}
\end{center}
\vspace{-10pt}
\end{table*}

\vspace{-3pt}\section{Method}
\vspace{-3pt}

\subsection{Overview of \model{}}
\vspace{-3pt}
In this section, we present the details of \model{}, which involves the following algorithm steps, including world model learning, behavior learning, and environment interaction: 
\begin{enumerate}[leftmargin=*]
    \vspace{-5pt}
    \item \textit{Affordance map computation} (Sec. \ref{sec:AffordanceMapAnnotation}): We employ a sliding bounding box to scan individual images and execute continuous zoom-ins inside the bounding box, simulating consecutive video frames that correspond to long-distance state transitions. We then create affordance maps by assessing the relevance of the fake video clips to task-specific goals expressed in text using the established MineCLIP reward model~\citep{fan2022minedojo}.
    \vspace{-2pt}\item \textit{Rapid affordance map generation} (\secref{sec:affordance_map_gen_with_unet}): Given that affordance maps will be frequently used in subsequent Step 5 to evaluate the necessities for jumpy state transitions, we train a U-Net module to approximate the affordance maps annotated in Step 1 for the sake of efficiency.
    \vspace{-2pt}\item \textit{World model training} (\secref{sec:repre_learn}): We train the world model to capture short- and long-term state transitions, using replay data with high responses from the affordance map. Each trajectory from the buffer includes pairs of samples from both adjacent time steps and long-distance intervals.
    \vspace{-2pt}\item \textit{Behavior learning} (\secref{sec:behav_learn}): We perform an actor-critic algorithm to optimize the agent's policy based on a finite sequence of long short-term imaginations generated by the world model. 
    \vspace{-2pt}\item \textit{Data update}: We apply the agent to interact with the environment and gather new data. Next, we leverage the generated affordance map to efficiently filter sample pairs suitable for long-term modeling, incorporating both short- and long-term sample pairs to update the replay buffer.
    \item Iterate Steps 3--5.
    \vspace{-5pt}
\end{enumerate}
Below, we discuss each training step in detail. The full algorithm can be found in \appref{sec:algo}.

% , first, we annotate affordance maps with MineCLIP and generate them using multimodal U-Net~(\secref{sec:affordance_map_gen}).
% %
% For representation learning, we introduce the world model with long short-term imagination~(\secref{sec:repre_learn}). 
% %
% For behavior learning, we adopt an actor-critic algorithm in long short-term imaginations of the world model~(\secref{sec:behav_learn}), encouraging the agent to focus on behaviors with long-term values. 

\vspace{-3pt}
\subsection{Affordance Map and Intrinsic Reward}
\label{sec:affordance_map_gen}
\vspace{-3pt}

We generate affordance maps using visual observations and textual task definitions to improve the sample efficiency of model-based reinforcement learning in open-world tasks. The core idea is to direct the agent's attention to task-relevant areas of the visual observation, leading to higher exploration efficiency. 
Let $\mathcal{M}_{o_t, I}(w, h)$ be the affordance map that represents the potential exploration value at pixel position $ (w, h) $ on the image observation $ o_t $, given textual instruction $I$ (\textit{e.g.}, ``cut a tree''). 
The affordance map highlights the relevance between regions of the observation and the task description, serving as a spatial prior that effectively directs the agent's exploration toward areas of interest.

\vspace{-3pt}
\subsubsection{Affordance Map Computation via Virtual Exploration}
\label{sec:AffordanceMapAnnotation}
\vspace{-3pt}

\begin{figure*}[t]
\vspace{-15pt}
    \centering
    \subfigure[Affordance map calculation]
    {\includegraphics[height=0.36\textwidth]{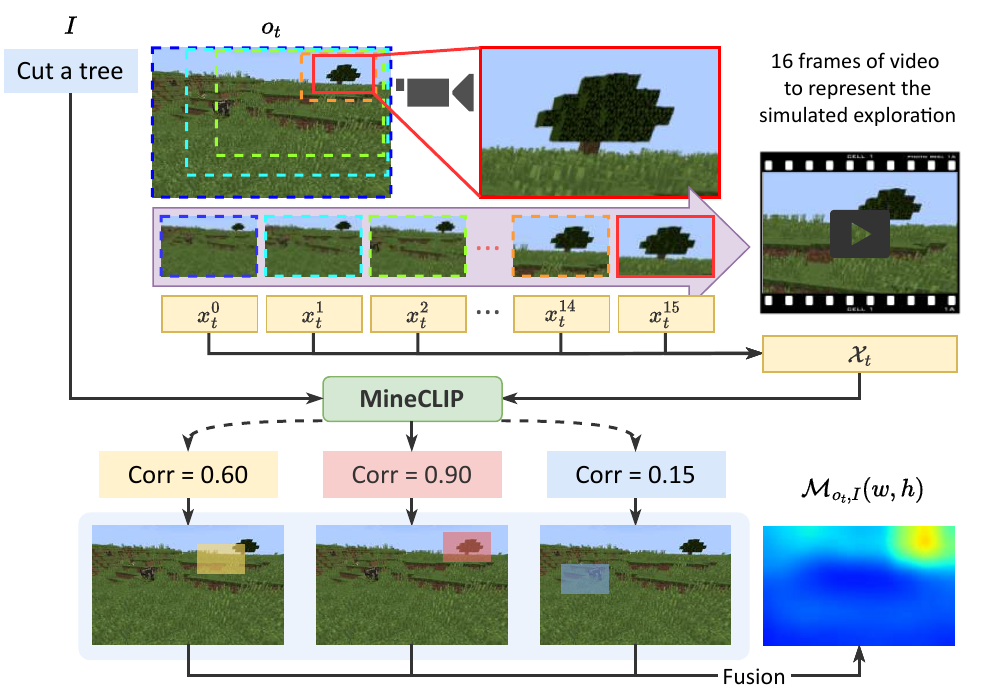}
        \label{fig:afford_compute}
    }
    \subfigure[Rapid affordance map generation]{\includegraphics[height=0.38\textwidth]{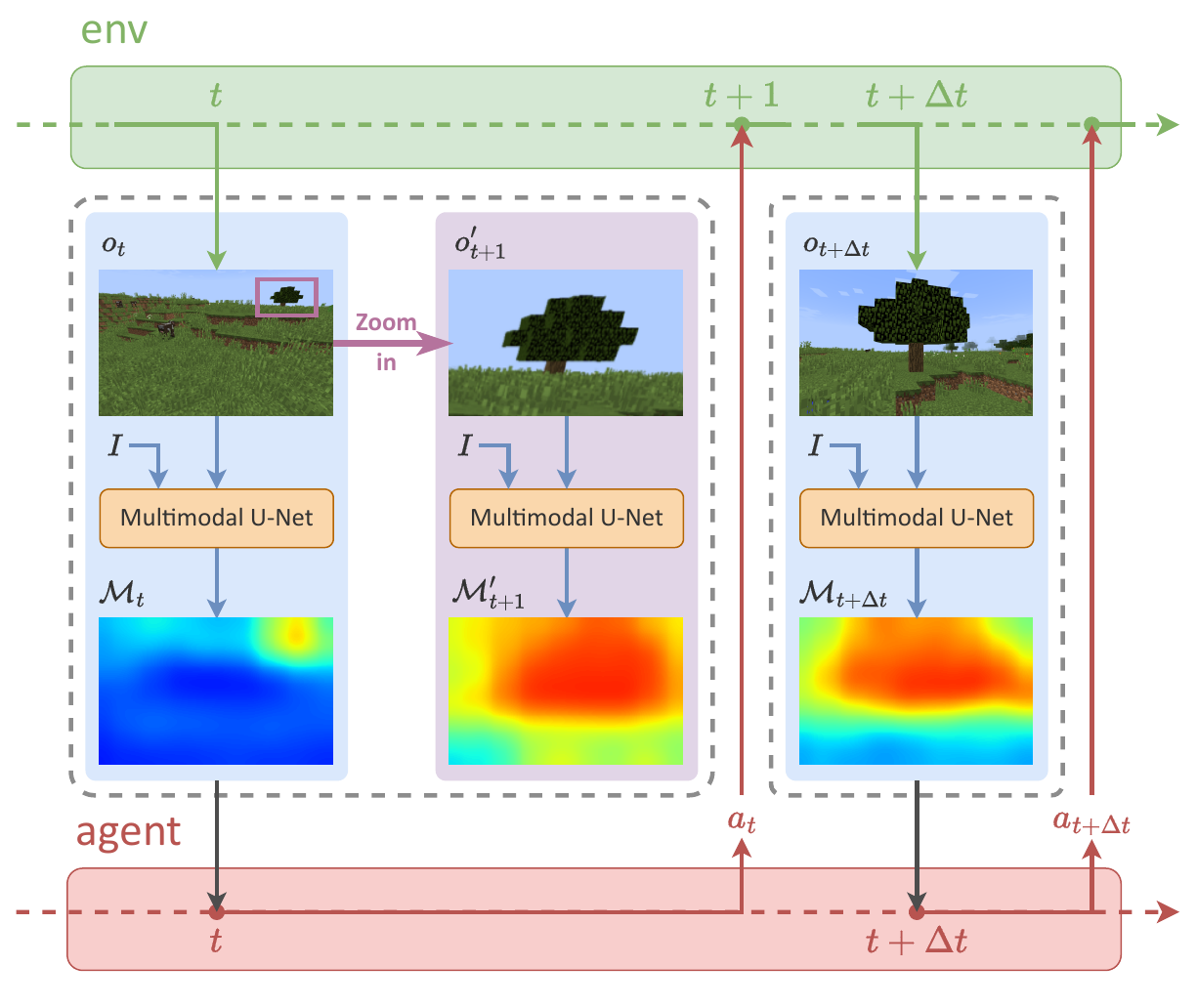}
    \label{fig:unet_gen}
    }
    \vspace{-12pt}
    \caption{The two steps for on-the-fly affordance map estimation: (a) Simulate exploration via image zoom-in and calculate the task-correlation scores of the virtual explorations using MineCLIP. (b) Learn to generate affordance maps more efficiently using a multimodal U-Net.}
    \vspace{-10pt}
    \label{fig:some_results}
\end{figure*}

To create the affordance map, as shown in \figref{fig:afford_compute}, we simulate and evaluate the agent's exploration without relying on real successful trajectories.
Concretely, we first adopt a random agent to interact with task-specific environments for data collection.
Starting with the agent's observation $ o_t $ at time step $t$, we use a sliding bounding box with dimensions scaled to $15\%$ of the observation's width and height to traverse the entire observation from left to right and top to bottom. 
The sliding bounding box moves horizontally and vertically in $9$ steps, respectively, covering every potential region in both dimensions.
For each position on the sliding bounding box of the observation $ o_t $, we crop $16$ images from $ o_t $.
These cropped images narrow the field of view to focus on the region and are resized back to the original image dimensions.
These resized images are denoted as $ x_{t}^{k} $ (where $ 0 \leq k < 16 $). The ordered set $ \mathcal{X}_{t} = [ x_{t}^{k} \mid k = 0, 1, \ldots, 15 ] $ represents a sequence of $16$ frames simulating the visual transition as the agent moves towards the position specified by the current sliding bounding box.
Subsequently, we employ the MineCLIP model\footnote{\reb{MineCLIP~\citep{fan2022minedojo} pretrains a video-language representation using Minecraft videos, enabling it to compute the correlation between a text string and a $16$-frame video segment.}} to calculate the correlation between the $\mathcal{X}_{t}$ of images, simulating the virtual exploration process, and the task description $I$. 
In this way, we quantify the affordance value of the sliding bounding box, indicating the potential exploration interest of the area.
%Detailedly, the region corresponding to the sliding window is assigned the correlation score, and other areas are set to zero. 
%
% These affordance maps are stored in the discrete set $\mathcal{S}_{\mathcal{M}, t}$.
After calculating the correlation score for each sliding bounding box, we fuse these values to obtain a smooth affordance map $\mathcal{M}_{o_t, I}$. For pixels that are covered by multiple sliding bounding boxes due to overlapping regions, the integrated affordance value is obtained by averaging the values from all the overlapping windows.

\vspace{-3pt}
\subsubsection{Multimodal U-Net for Rapid Affordance Map Generation}
\label{sec:affordance_map_gen_with_unet}
\vspace{-3pt}

% The generation of affordance maps through the method discussed previously, which utilizes virtual exploration and a pre-trained video-text alignment model, is computationally intensive due to the need for extensive window traversal and calculations for each window. This process is not only time-consuming but also challenging to balance efficiency and performance, making real-time affordance map generation difficult. To address this limitation, we adopt a self-supervised learning approach, utilizing the previously annotated affordance maps as supervisory signals to train a model capable of rapidly generating exploration affordance maps based on images and their corresponding instructions, catering to the demands of real-time applications.

The annotation of affordance maps, as previously described, involves extensive window traversal and computations for each window position using a pre-trained video-text alignment model.
This method is computationally demanding and time-consuming, making real-time applications challenging. 
To address this issue, we first use a random agent to interact with the environment for data collection. 
Next, we annotate the affordance maps for the collected images using the aforementioned method based on virtual exploration.
We gather a dataset of tuples ($o_t,I,\mathcal{M}_{o_t, I}$) and use it to train a multimodal U-Net based on Swin-Unet~\citep{swinunet}.
To handle multimodal inputs, we extract text features from the language instructions and image features from the downsampling process of Swin-Unet, and fuse them with multi-head attention.
We present architecture details in \figref{fig:multimodal_unet} in the appendix. 
In this way, with the pretrained multimodal U-Net, we can efficiently generate affordance maps at each time step using visual observations and language instructions.

% Our enhanced Multimodal U-Net architecture significantly builds upon the Swin-Unet~\citep{swinunet} framework by integrating multimodal capabilities, crucial for merging textual task directives with visual information. This synthesis is primarily facilitated through the Text Image Attention (TIA) module, which employs a multi-head attention mechanism to intricately weave deep-level image features with textual descriptions. This process occurs within the U-Net's deeper layers before transitioning into the bridge layer, ensuring the annotated heatmaps accurately depict the relationship between different image regions and the corresponding textual instructions. Additionally, the architecture incorporates the Grouped multi-axis Hadamard Product Attention (GHPA) and the Group Aggregation Bridge (GAB) modules, derived from the Swin-Unet design. The GHPA module is instrumental in refining the encoder's capacity to highlight critical features, whereas the GAB module facilitates the smooth amalgamation of features across varying network levels, enhancing the overall robustness of feature extraction and integration within our model.

% \begin{figure}[t]
%     \centering
%     \includegraphics[width=\textwidth]{fig/affordance_generation.pdf}
%     \vspace{-15pt}
%     \caption{
%     Affordance map annotation 
%      }
%     % The overall transfer learning pipeline of CoWorld involves two model-based RL agents and three training stages. Please refer to the text in \cref{sec:overall_pipline} for detailed descriptions.
%     \label{fig:afford_gen}
%     \vspace{-12pt}
% \end{figure}

\vspace{-3pt}
\subsubsection{Affordance-Driven Intrinsic Reward}
\label{sec:final_reward}
\vspace{-3pt}

To leverage the task-relevant prior knowledge presented by the affordance map for efficient exploration in the open world, we introduce the following intrinsic reward function:
\begin{equation}
\label{eq:intrinsic_reward}
r^\text{intr}_t = \frac{1}{WH} \sum_{w=1}^W \sum_{h=1}^H \mathcal{M}_{o_t,I}(w, h) \cdot \mathcal{G}(w, h),
\end{equation}
where $ W $ and $ H $ denote the width and height of the visual observation. 
$\mathcal{G}$ represents a Gaussian matrix with dimensions matching those of the affordance map.
It corresponds to a 2D Gaussian distribution, with its peak located at the center of the affordance map.
The values in the matrix are determined by standard deviations $(\sigma_x, \sigma_y)$, while the mean is uniformly set to $1$ across the entire matrix. We present visualizations of $\mathcal{G}$ and conduct hyperparameter analyses on $(\sigma_x, \sigma_y)$ in Appendix \ref{sec:hyper}.
The intuition behind this design is to encourage the agent to move toward the target.

Overall, the agent receives a composite reward consisting of the episodic sparse reward from the environment, the reward from MineCLIP~\citep{fan2022minedojo}, and the intrinsic reward from the affordance map: $r_t=r^\text{env}_t+r_t^{\text{MineCLIP}}+\alpha r_t^\text{intr}$,
% \begin{equation}
% \label{eq:final_reward}
% r_t = r^\text{env}_t+r_t^{\text{MineCLIP}}+\alpha r_t^\text{intr},
% \end{equation} 
where $\alpha$ is a hyperparameter.
In contrast to the MineCLIP reward, which relies on the agent's past performance, our affordance-driven intrinsic reward emphasizes long-term values derived from future virtual exploration.
It encourages the agent to adjust the policy to pursue task-related targets when they appear in its view, ensuring these targets are centrally positioned in future visual observations to maximize this reward function. 

% Simultaneously, we can determine from the distribution of the affordance map whether there are task-related targets within the observed image that are at a certain distance from the agent. 
%

\vspace{-3pt}
\subsection{Long Short-Term World Model}
\label{sec:repre_learn}
\vspace{-3pt}

\subsubsection{Learning Jumping Flags}
\vspace{-3pt}

In \model{}, the world model is customized for long-term and short-term state transitions. It decides which type of transition to adopt based on the current state and predicts the next state with the selected transition branch. 
To facilitate the switch between long-term and short-term state transitions, we introduce a jumping flag $j_t$, which indicates whether a jumpy transition or long-term state transition, should be adopted at time step $t$. 
When a distant task-related target appears in the agent’s observation, which can be reflected by a higher kurtosis in the affordance map, a jumpy transition allows the agent to imagine the future state of approaching the target.
To this end, we define relative kurtosis $K_r$ which measures whether there are significantly higher target areas than the surrounding areas in the affordance map, and absolute kurtosis $K_a$ represents the confidence level of target presence in that area. 
Formally, 
\begin{equation}
\begin{aligned}
K_r &= \frac{1}{WH}\sum_{w=1}^W \sum_{h=1}^H \left[\left(\frac{\mathcal{M}_{o,I}(w, h) - \mathrm{mean}(\mathcal{M}_{o,I})}{\mathrm{std}(\mathcal{M}_{o,I})}\right)^4\right], \\
K_a &= \mathrm{max}(\mathcal{M}_{o,I}) - \mathrm{mean}(\mathcal{M}_{o,I}).
\end{aligned}
\end{equation}
To normalize the relative kurtosis, we apply the sigmoid function to it,
and then multiply it by the absolute kurtosis to calculate the jumping probability:
\begin{equation}
\label{eq:jump_prob}
    P_{\text{jump}} = \mathrm{sigmoid}(K_r) \times K_a.
\end{equation}
The jumping probability measures the confidence in the presence of task-relevant targets far from the agent in the visual observation. 
To determine whether to employ long-term state transition, we use a dynamic threshold, which is the mean of the collected jumping probabilities at each time step, plus one standard deviation. \reb{For a detailed explanation, please refer to \ref{sec:env_interact}.
}
If $P_{\text{jump}}$ exceeds this threshold, the jump flag $j_t$ is \texttt{True} and the agent switches to jumpy state transitions in the imagination phase.

\vspace{-3pt}
\subsubsection{Learning Jumpy State Transitions}
\vspace{-3pt}

\begin{figure}[t]
\vspace{-15pt}
    \centering
    \includegraphics[width=1.0\textwidth]{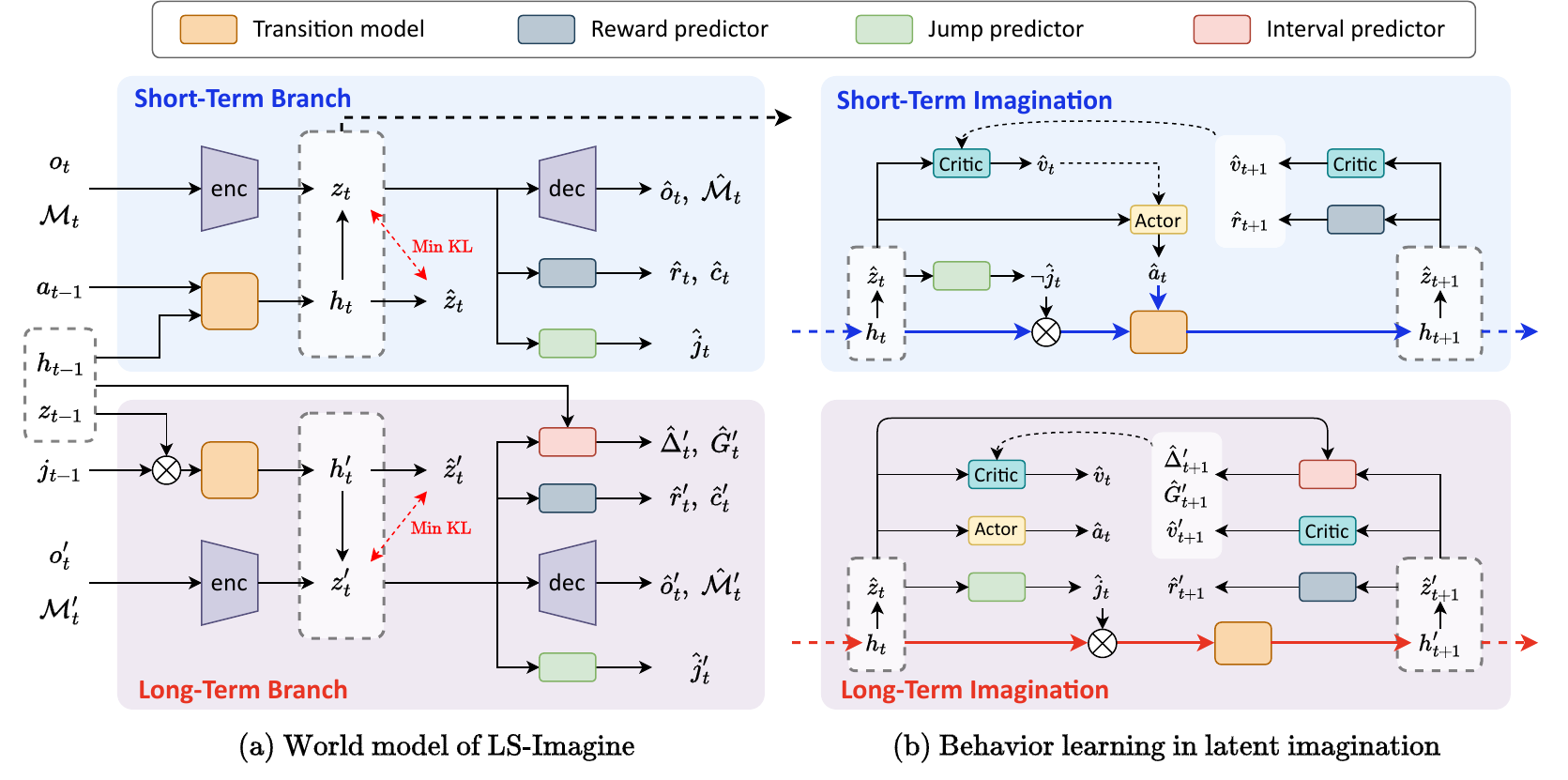}
    \vspace{-20pt}
    \caption{
    The overall architecture of the world model and the behavior learning process.
     }
    \label{fig:wm_learn}
    \vspace{-10pt}
\end{figure}

In \model{}, the state transition model includes both short-term and long-term branches. As shown in \figref{fig:wm_learn} (a), the short-term transition model integrates the previous deterministic recurrent state $ h_{t-1} $, stochastic state $ z_{t-1} $, and action $ a_{t-1}$ to adopt the single-step transition. 
In contrast, the long-term branch simulates jumpy state transitions toward the target.
\reb{It is important to clarify that the index $t$ does not denote the time step during real environmental interactions but instead represents the positional order of states within the imagination sequence.}
The overall world model of \model{} is primarily based on DreamerV3~\citep{hafner2023dreamerv3}, with novel components specifically designed to capture jumpy state transitions:
\eq{
\label{eq:world_model}
\begin{alignedat}{3}
    &\text{Short-term transition model:} &\quad h_t &= f_\phi(h_{t-1}, z_{t-1}, a_{t-1}) \vphantom{\hat{\Delta}_{t}^{\prime}} \\
    &\text{Long-term transition model:}  &\quad {h}_{t}^{\prime} &= f_\phi(h_{t-1}, z_{t-1}) \vphantom{\hat{\Delta}_{t}^{\prime}} \\
    &\text{Encoder:} &\quad z_t &\sim  q_\phi(z_t \mid h_t, o_t, \mathcal{M}_t) \vphantom{\hat{\Delta}_{t}^{\prime}} \\
    &\text{Dynamics predictor:} &\quad \hat{z}_{t} &\sim p_{\phi}(\hat{z}_{t} \mid h_{t}) \vphantom{\hat{\Delta}_{t}^{\prime}} \\ 
    &\text{Reward predictor:} &\quad \hat{r}_{t}, \hat{c}_{t} &\sim p_\phi(\hat{r}_t, \hat{c}_{t} \mid h_t, z_t) \vphantom{\hat{\Delta}_{t}^{\prime}} \\
    &\text{Decoder:} &\quad \hat{o}_t, \hat{\mathcal{M}}_t &\sim  p_\phi(\hat{o}_t, \hat{\mathcal{M}}_t \mid h_t, z_t) \vphantom{\hat{\Delta}_{t}^{\prime}} \\
    &\text{Jump predictor:} &\quad \hat{j}_{t} &\sim p_\phi( \hat{j}_{t} \mid h_t, z_t) \vphantom{\hat{\Delta}_{t}^{\prime}} \\
    &\text{Interval predictor:} &\quad {\hat{\Delta}_{t}}^{\prime}, {\hat{G}_{t}}^{\prime} &\sim p_\phi({\hat{\Delta}_{t}}^{\prime}, {\hat{G}_{t}}^{\prime} \mid h_{t-1}, z_{t-1}, {h}^{\prime}_{t}, {z}^{\prime}_{t})
\end{alignedat}.
}
At time step $t$, we feed the recurrent state $h_t $, the observation $o_t $, and the affordance map $ \mathcal{M}_t $ into the encoder to obtain posterior state $z_t$. We also use the affordance map as an input of the encoder, which serves as the goal-conditioned prior guidance to the agent.
Notably, the prediction of prior state $\hat{z}_t$ does not involve the current observation or affordance map, relying solely on historical information. 
We use $(h_t,z_t)$ to reconstruct the visual observation $ \hat{o}_t$ and the affordance map $\hat{\mathcal{M}}_t $, and predict the reward $ \hat{r}_t $, episode continuation flag $ \hat{c}_t$, and jumping flag $\hat{j}_t$.
For long-term state transitions, we use an interval predictor to estimate the expected number of interaction steps $\hat{\Delta}_t^\prime$ required to transition from the pre-jump state $ (h_{t-1}, z_{t-1}) $ to the post-jump state $ ({h}^{\prime}_{t}, {z}^{\prime}_{t}) $, along with the expected cumulative reward $\hat{G}_t^\prime$ that the agent may receive during this time interval. 
\reb{We detail the approach to annotate $\Delta_t^\prime$ and $G_t^\prime$ using the real interaction data in \appref{sec:env_interact}.}

% \yb{todo: To train the world model, we interact with the environment to collect short-term tuples $\mathcal{D}_t= (o_t,a_t,r_t,c_t, \mathcal{M}_t, j_t,\Delta_t, G_t)$ and long-term tuples $\mathcal{D}^{\prime}_t=(o_t^{\prime},a_t^{\prime},{r}_t^{\prime}, \mathcal{M}^{\prime}_t, c_t^{\prime},j_t^{\prime},\Delta_t^{\prime}, G_t^{\prime})$ using the current policy.}
\reb{We collect short-term tuples $\mathcal{D}_t= (o_t,a_t, \mathcal{M}_t, r_t,c_t,j_t,\Delta_t, G_t)$ from each interaction with the environment using the current policy. When observing $j_t = 1$, we additionally construct long-term tuples $\mathcal{D}^{\prime}_{t+1}=(o_{t+1}^{\prime},a_{t+1}^{\prime}, \mathcal{M}^{\prime}_{t+1}, {r}_{t+1}^{\prime},c_{t+1}^{\prime},j_{t+1}^{\prime},\Delta_{t+1}^{\prime}, G_{t+1}^{\prime})$ based on $\mathcal{D}_t$.}
More details for this process can be found in \appref{sec:env_interact}.
%
% the calculation methods for the affordance map $ {\mathcal{M}}^{\prime}_{t} $, intrinsic reward $ {r_{t}^{i}}^{\prime} $, and jumpy transition flag $ {j_{t}}^{\prime} $ are the same as those before the jumpy transition. 
%
During representation learning, we sample short-term tuples $\{\mathcal{D}_t\}_{t=1}^{T}$ and the long-term tuples following jumpy transitions \reb{$\{\mathcal{D}_{t+1}^{\prime}\}_{t \in \mathcal{T}}$} from the replay buffer $\mathcal{B}$, where $\mathcal{T}$ denotes the set of time steps at which long-term state transitions are required. The loss functions for each component of the short-term and long-term world model branch are as follows:
\begin{equation}
    \begin{aligned}
    &\text{Short-term branch:} \quad \left\{
    \begin{aligned}
        \mathcal{L}_{\text{dyn}} &\doteq  \max \left(1, \operatorname{KL}\left[\operatorname{sg}\left(q_{\phi}\left(z_{t} \mid h_{t}, o_{t}, \mathcal{M}_{t}\right)\right) \parallel  p_{\phi}\left(z_{t} \mid h_{t}\right)\right]\right) \\
        \mathcal{L}_{\text {enc}} &\doteq  \max \left(1, \operatorname{KL}\left[ q_{\phi}\left(z_{t} \mid h_{t}, o_{t}, \mathcal{M}_{t}\right) \parallel \operatorname{sg} \left( p_{\phi}\left(z_{t} \mid h_{t}\right)\right)\right]\right) \\
        \mathcal{L}_{\text {dec}} &\doteq - \operatorname{ln} p_{\phi} \left(o_t,\mathcal{M}_t \mid h_t, z_t \right) \\
        \mathcal{L}_{\text {pred}} &\doteq - \operatorname{ln} p_{\phi} \left({r}_{t}, {c}_{t} \mid h_t, z_t \right) - \operatorname{ln} p_{\phi} \left( {j}_{t} \mid h_t, z_t \right)
    \end{aligned}
    \right. \\
    \end{aligned}.
\end{equation}

\begin{equation}
    \begin{aligned}
    &\text{Long-term branch:} \quad \left\{
    \begin{aligned}
        \mathcal{L}^{\prime}_{\text {dyn}} &\doteq \max \left(1, \operatorname{KL}\left[\operatorname{sg}\left(q_{\phi}\left({z}^{\prime}_{t} \mid {h}^{\prime}_{t}, o^{\prime}_{t}, {\mathcal{M}}^{\prime}_{t}\right)\right) \parallel  p_{\phi}\left({z}^{\prime}_{t} \mid {h}^{\prime}_{t}\right)\right]\right) \\
        \mathcal{L}^{\prime}_{\text {enc}} &\doteq \max \left(1, \operatorname{KL}\left[ q_{\phi}\left({z}^{\prime}_{t} \mid {h}^{\prime}_{t}, o^{\prime}_{t}, {\mathcal{M}}^{\prime}_{t}\right) \parallel \operatorname{sg} \left( p_{\phi}\left({z}^{\prime}_{t} \mid {h}^{\prime}_{t}\right)\right)\right]\right) \\
        \mathcal{L}^{\prime}_{\text {dec}} &\doteq - \operatorname{ln} p_{\phi} \left(o^{\prime}_{t},{\mathcal{M}}^{\prime}_{t} \mid {h}^{\prime}_{t}, {z}^{\prime}_{t} \right) \\
        \mathcal{L}^{\prime}_{\text {pred}} &\doteq - \operatorname{ln} p_{\phi} ({r}^{\prime}_{t}, {c}^{\prime}_{t} \mid {h}^{\prime}_{t}, {z}^{\prime}_{t} ) - \operatorname{ln} p_{\phi} ({{j}}^{\prime}_{t} \mid {h}^{\prime}_{t}, {z}^{\prime}_{t} ) \\
        \mathcal{L}^{\prime}_{\text {int}} &\doteq - \operatorname{ln} p_{\phi} \left({\Delta}^{\prime}_{t}, G_{t}^{\prime} \mid h_{t-1}, z_{t-1}, {h}^{\prime}_{t}, {z}^{\prime}_{t} \right)
    \end{aligned}
    \right.
    \end{aligned}.
\end{equation}

We can optimize the world model $\mathcal{W}_{\phi}$ by minimizing over replay buffer $\mathcal{B}$:
\begin{equation}
    \begin{aligned}
    \mathcal{L} \doteq \mathbb{E}\Big[  &{\textstyle \sum_{ \{\mathcal{D}_t\}_{t=1}^{T}}} \left( \beta_{\text{dyn}} \mathcal{L}_{\text{dyn}} + \beta_{\text{enc}} \mathcal{L}_{\text{enc}} + \beta_{\text{pred}} \left( \mathcal{L}_{\text{dec}} + \mathcal{L}_{\text{pred}} \right)\right) + \\
    \beta_{\text{long}}&{\textstyle \sum_{{\{  \reb{\mathcal{D}^{\prime}_{t+1}}}  \}_{t\in \mathcal{T}} }} \left( \beta_{\text{dyn}} \mathcal{L}^{\prime}_{\text{dyn}} + \beta_{\text{enc}} \mathcal{L}^{\prime}_{\text{enc}} + \beta_{\text{pred}} \left( \mathcal{L}^{\prime}_{\text{dec}} + \mathcal{L}^{\prime}_{\text{pred}} + \mathcal{L}^{\prime}_{\text{int}} \right)\right)  \Big].
    \end{aligned}
    \label{eq:dyn_wm_loss}
\end{equation}
% where $\beta_{\text{dyn}}$, $\beta_{\text{enc}}$, $\beta_{\text{pred}}$, and $\beta_{\text{long}}$ are hyperparameters, and $\beta_{\text{long}}$ governs the importance of long-term modeling.
% \ljj{t+1?}
%%%%%%%%%%%%%%%%%%%%%%%%%%%%%%%%%%%%%%%%%%%%%%%%%%%
\vspace{-5pt}
\subsection{Behavior Learning over Mixed Long Short-Term Imaginations}
\label{sec:behav_learn}
\vspace{-3pt}

As shown in \figref{fig:wm_learn} (b), \model{} employs an actor-critic algorithm to learn behavior from the latent state sequences predicted by the world model. 
The goal of the actor is to optimize the policy to maximize the discounted cumulative reward $R_t$, while the role of the critic is to estimate the discounted cumulative rewards using the current policy for each state \reb{$\hat{s}_t \doteq \{h_t,\hat{z}_t\}$}:
\begin{equation}
    \text{Actor:} \quad \hat{a}_t \sim \pi_{\theta} \left( \hat{a}_t \mid \hat{s}_t \right), \quad \text{Critic:} \quad v_{\psi} \left( \hat{R}_t \mid \hat{s}_t \right).
\end{equation}
Starting from the initial state encoded from the sampled observation and the affordance map, we dynamically select either the long-term transition model or the short-term transition model to predict subsequent states based on the jumping flag $\hat{j}_t$.
For the long short-term imagination sequence $\{(\hat{s}_t, \hat{a}_t)\}_{t=1}^L$ with an imagination horizon of $L$, we predict reward sequence $\hat{r}_{1:L}$ and the continuation flag sequence $\hat{c}_{1:L}$ through the reward predictor. 
\reb{Similar to \eqref{eq:world_model}, the index $t$ does not represent the time step in the environment, but rather the positional order of the states in the imagination sequence. Specifically, starting from state $\hat{s}_t$, any subsequent state obtained via either a short-term transition or a long-term transition is indexed sequentially as $t+1$.}

For jumpy states predicted by long-term imagination, the interval predictor estimates \reb{(i) the number of steps $\hat{\Delta}_{t}$ from $\hat{s}_{t-1}$ to $\hat{s}_{t}$ and (ii) the potential discounted cumulative reward $\hat{G}_{t}$ over the time interval of $\hat{\Delta}_{t}$.
Otherwise, for states obtained via short-term imagination, which correspond to single-step transitions in the environment, we set $\hat{\Delta}_{t}=1$ and $\hat{G}_{t}=\hat{r}_{t}$. 
Consequently, within one imagination episode, we obtain a sequence of step intervals $\hat{\Delta}_{2:L}$ and a sequence of predicted rewards $\hat{G}_{2:L}$ between consecutive imagination states.}

We employ a modified bootstrapped $\lambda$-returns that considers both long-term and short-term imaginations to calculate the discounted cumulative rewards for each state:
%
% When $t=L$, the return $R^\lambda_t$ is defined as the terminal value $v_\psi(\hat{s}_L)$. For  $t < L$ , the return $ R^\lambda_t $ is given by:
% \begin{equation}
% \label{eq:value_compute}
% R_{t}^{\lambda} \doteq \left\{
% \begin{array}{ll}
% \hat{c}_{t} \left(\hat{G}_{t+1} + \gamma^{\hat{\Delta}_{t+1}} \left[ (1-\lambda) v_{\psi} (\hat{s}_{t+1}^{\prime}) + \lambda R_{t+1}^{\lambda \prime}\right] \right) & \text{if } \hat{j}_t=\texttt{True} \\
% \hat{c}_{t} \left(\hat{r}_{t+1} + \gamma \left[(1-\lambda) v_{\psi} (\hat{s}_{t+1}) + \lambda R_{t+1}^{\lambda} \right] \right) & \text{if } \hat{j}_t=\texttt{False}
% \end{array}
% \right.
% .
% \end{equation}
\begin{equation}
\label{eq:value_compute}
R_{t}^{\lambda} \doteq \left\{
\begin{array}{ll}
\hat{c}_{t} \{\hat{G}_{t+1} + \gamma^{\hat{\Delta}_{t+1}} \left[ (1-\lambda) v_{\psi} (\hat{s}_{t+1}) + \lambda R_{t+1}^{\lambda} \right] \} & \text{if } t < L \\
v_{\psi} (\hat{s}_{L}) & \text{if } t = L
\end{array}
\right.
.
\end{equation}

% Where $\hat{s}^{\prime}_{t+1}$ is post-jumping state, and $R_{t+1}^{\lambda \prime}$ is actually approximated by $R_{t+\hat{\Delta}_t}^{\lambda}$. For simplicity, we denote it as $\lambda R_{t+1}^{\lambda \prime}$.}
The critic uses the maximum likelihood loss to predict the distribution of the return estimates $R_{t}^{\lambda}$:
\begin{equation}
\label{eq:critic_loss}
\mathcal{L}(\psi) \doteq -\sum_{t=1}^{L} \ln p_{\psi}\left(R_{t}^{\lambda} \mid \hat{s}_t \right).
\end{equation}
Following DreamerV3~\citep{hafner2023dreamerv3}, we train the actor to maximize the return estimates $R_t^{\lambda}$. 
Notably, since long-term imagination does \textit{not} involve actions, we do not optimize the actor at time steps when jumpy state transitions are adopted.
\reb{
Therefore, unlike DreamerV3, we apply an additional factor of $ (1 - \hat{j}_t) $ to ignore updates at long-term imagination steps:
}
\begin{equation}
\label{eq:actor_loss}
\mathcal{L}(\theta) \doteq -\sum_{t=1}^{L} \mathrm{sg} \left[\left(1-\hat{j}_{t}\right)\frac{R_t^\lambda - v_\psi(\hat{s}_t)}{\max(1, S)} \right] \log \pi_\theta(\hat{a}_t \mid \hat{s_t}) + \eta \, \mathrm{H} \left[ \pi_\theta(\hat{a}_t \mid \hat{s_t}) \right].
\end{equation}

% To enhance robustness against outliers, we calculate the range between the $5^{\text{th}}$ and $95^{\text{th}}$ percentiles of the return distribution within the batch and then smooth this estimate with an exponential moving average:
% \begin{equation}
% S \doteq \operatorname{EMA}\left(\operatorname{Per}\left(R_{t}^{\lambda}, 95\right)-\operatorname{Per}\left(R_{t}^{\lambda}, 5\right), 0.99\right).
% \end{equation}

% \yb{Fig. 4: the imagination pipeline --- how to switch between the short-term and long-term imaginations}

\vspace{-3pt}
\section{Experiments}
\label{sec:experiment}
\vspace{-3pt}
% \subsection{Experimental Setups}
% \vspace{-3pt}

% \paragraph{Benchmark.} 
We explore \model{} on the challenging MineDojo~\citep{fan2022minedojo} benchmark on top of the popular Minecraft game, which is a comprehensive simulation platform with various open-ended tasks.
We use $5$ tasks, \ie \textit{harvest log in plains}, \textit{harvest water with bucket}, \textit{harvest sand}, \textit{shear sheep}, and \textit{mine iron ore}.
These tasks demand numerous steps to complete and present significant challenges for agent learning.
We adopt a binary reward that indicates whether the task was completed, along with the MineCLIP reward \citep{fan2022minedojo}.
% For long-term imagination of \model{}, we use zoom-in reward as the intrinsic reward for sample-efficient exploration.
Further details of the environmental setups are provided in \appref{sec:env_detail}.
Besides, we introduce the compared models in \appref{sec:compared_methods}.

\begin{figure}[t]
\vspace{-15pt}
    \centering
    \includegraphics[width=\textwidth]{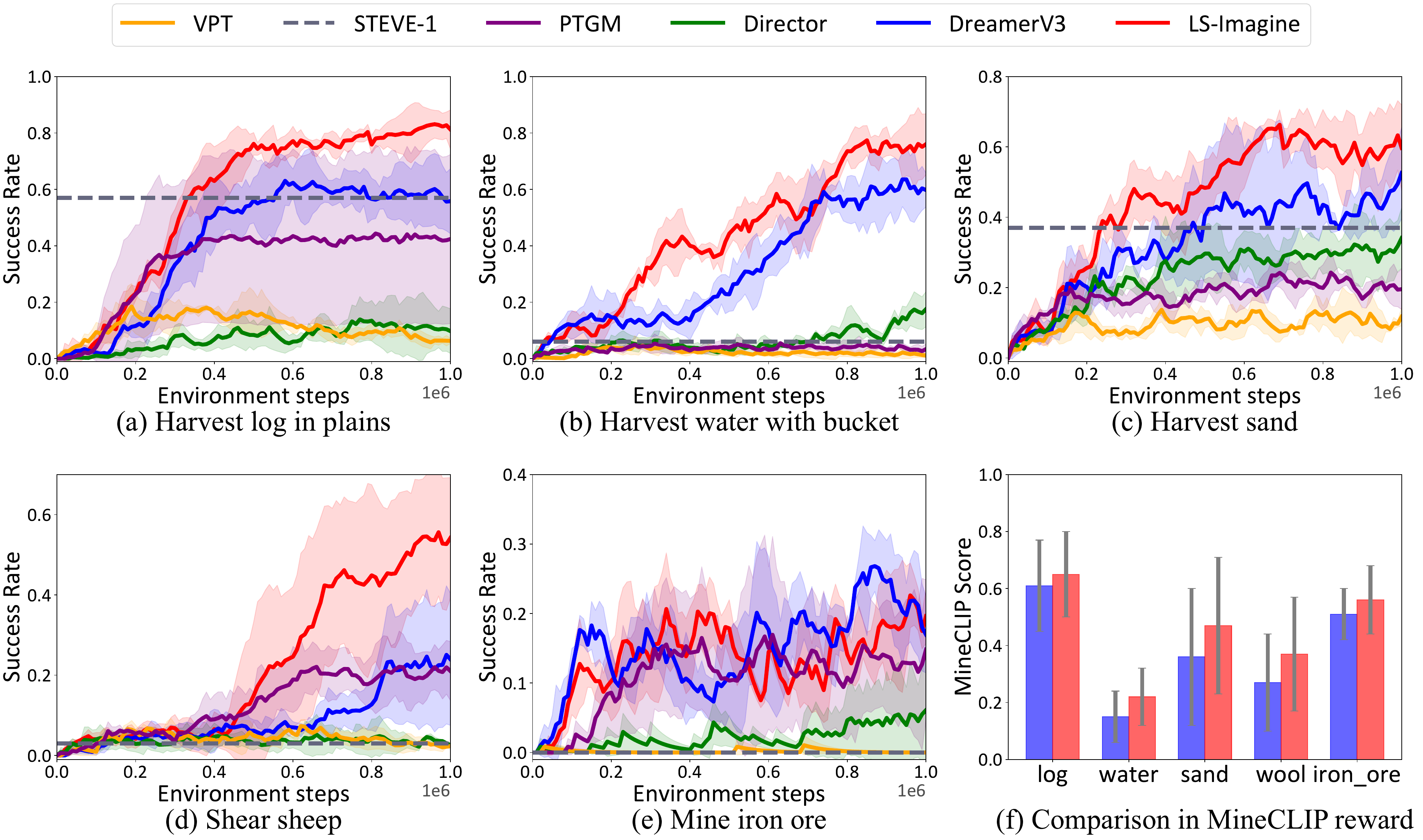}
    \vspace{-20pt}
    \caption{Comparison of \model{} against strong Minecraft agents, including \textit{DreamerV3}~\citep{hafner2023dreamerv3}, \textit{VPT}~\citep{baker2022video}, \textit{STEVE-1}~\citep{lifshitz2023steve}, \textit{PTGM}~\citep{yuan2024pre}, and \textit{Director}~\citep{hafner2022deep}. We present the numerical results in Table \ref{tab:minedojo_cmp_results} in the appendix.
    }
    % The overall transfer learning pipeline of CoWorld involves two model-based RL agents and three training stages. Please refer to the text in \cref{sec:overall_pipline} for detailed descriptions.
    \label{fig:baseline_comparion}
    \vspace{-10pt}
\end{figure}

\vspace{-8pt}
\paragraph{Implementation details.}
\label{sec:imple_detail}
We conduct our experiments on the MineDojo environment, where both visual observation and corresponding affordance maps are resized to $64 \times 64$ pixels. 
To generate accurate affordance maps, we collect $2{,}000$ images from the environment using a random agent under the current task instruction and generate a discrete set of ($o_t,I,\mathcal{M}_{o_t, I}$), which are then used to finetune the multimodal U-Net for $200$ epochs.
For tasks in the MineDojo benchmark, we train the agent for $1\times 10^6$ environment steps. Each training of \model{} takes approximately $23$ GB of VRAM and requires around $1.7$ days to complete on a single RTX 3090 GPU.
% The agent's performance was evaluated by tracking both the success rate of each episode and the number of steps required to achieve success. 
%
% We compared the performance of our \model{} with various baseline methods, focusing on sample efficiency and task performance across the training process. 
%

\vspace{-3pt}
\subsection{Main Comparison}
\vspace{-3pt}

% In the open-world environment of Minecraft, we selected three representative types of tasks for our experiments: sparse-target tasks (such as searching for and chopping trees on plains, finding water sources), dynamic-target tasks (such as shearing sheep), and mining tasks (such as mining iron ore). Details of these tasks are provided in \appref{sec:env_detail}. 

\begin{figure}[t]
\vspace{-5pt}
    \centering
    \includegraphics[width=\textwidth]{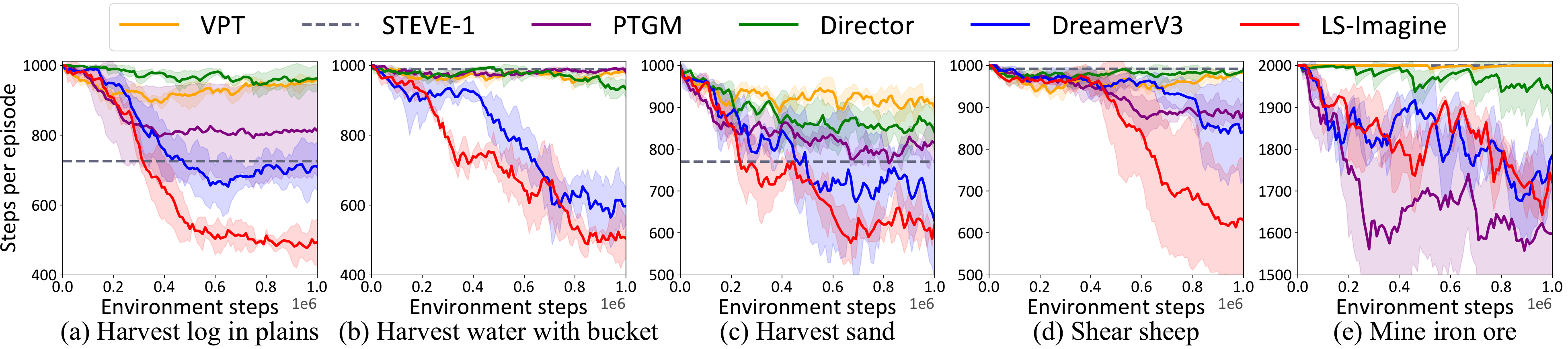}
    \vspace{-18pt}
    \caption{The number of steps per episode for task completion.}
    % The overall transfer learning pipeline of CoWorld involves two model-based RL agents and three training stages. Please refer to the text in \cref{sec:overall_pipline} for detailed descriptions.
    \label{fig:success_step_comp}
    \vspace{-10pt}
\end{figure}

We evaluate all the Minecraft agents in terms of success rate shown in \figref{fig:baseline_comparion} and per-episode steps shown in \figref{fig:success_step_comp}. 
%
% \figref{fig:baseline_comparion} and \figref{fig:success_step_comp} demonstrate the performance of \model{} and all the baselines across five different tasks. 
%
We find that \model{} significantly outperforms the compared models, particularly in scenarios where sparse targets are distributed in the task. 
In \figref{fig:baseline_comparion} (f), we showcase the MineCLIP values achieved by \model{} and DreamerV3.
Specifically, a sliding window of length $16$ is used to compute the local MineCLIP values for each segment. The mean value is then calculated from all sliding windows.
We can see that agents trained using our method achieve higher MineCLIP values within a single episode compared to DreamerV3. This suggests that \model{} facilitates quicker detection of task-relevant visual targets in open-world environments.

%
% Additionally, the parallel way increases the proportion of imagination sequences that are relatively close to the target. 

% It makes the behavior learning of the agent less prone to exploit the jumpy state transition effectively.

\begin{figure*}[t]
    \centering
    \subfigure[Visualization of long short-term imaginations]{\includegraphics[height=0.26\textwidth]{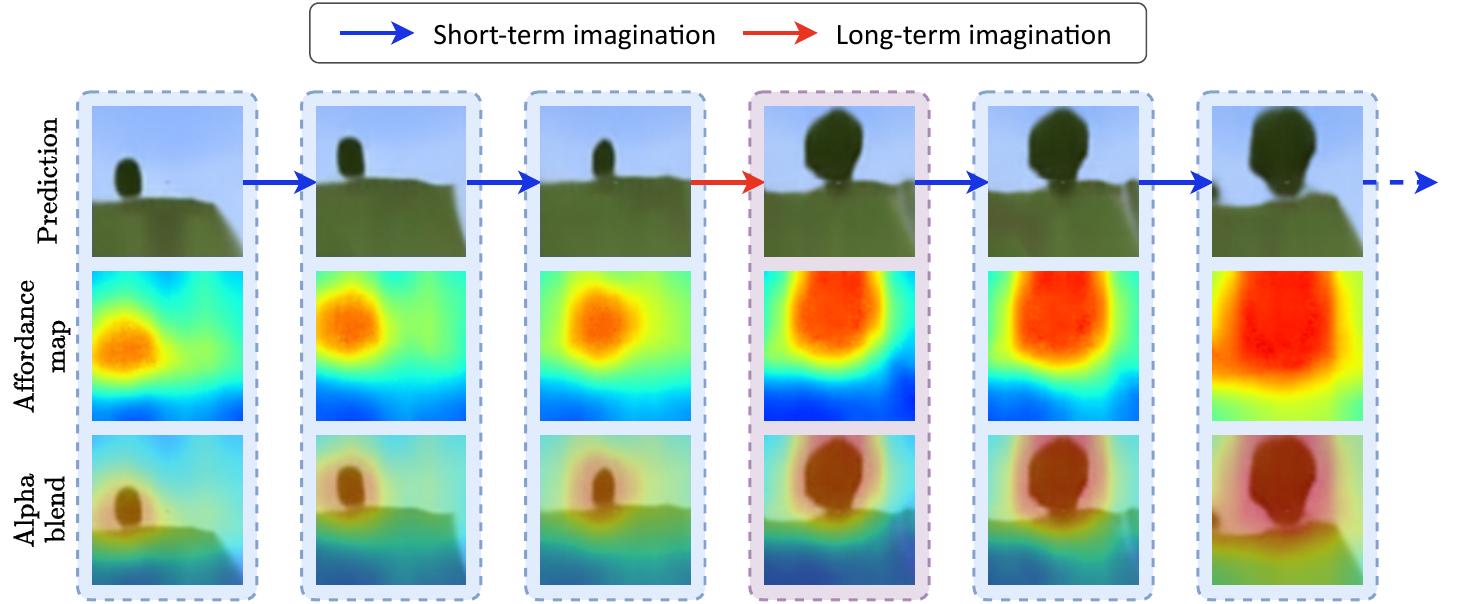}
    \label{fig:imagination_sequence}
    }
    \subfigure[Series vs. parallel connections]{\includegraphics[height=0.26\textwidth]{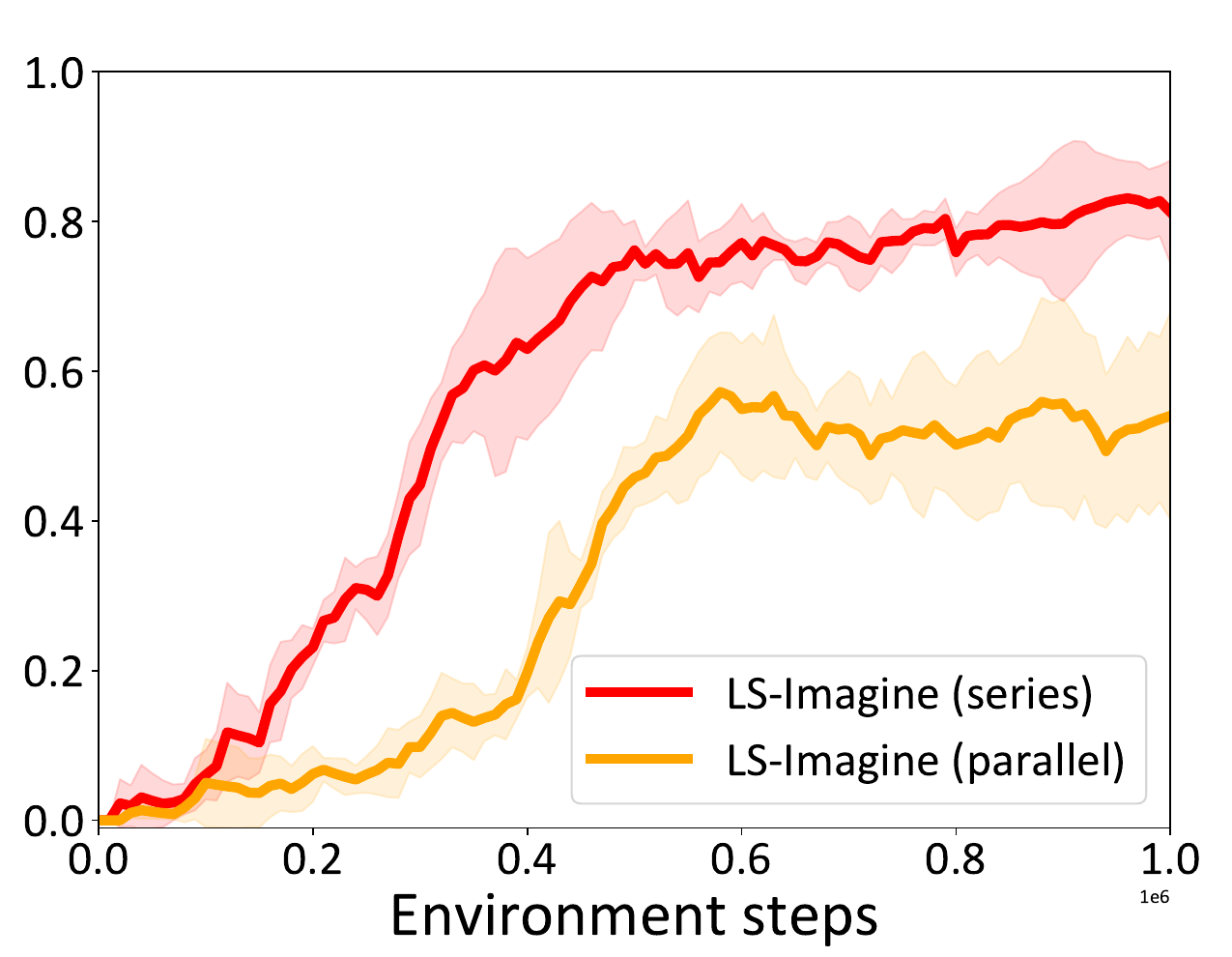}
        \label{fig:sequencial_and_parallel}
    }
    \vspace{-10pt}
    \caption{(a) Visualization of long short-term imaginations and (b) a further discussion on possible architecture designs of Series and Parallel connections of these two imagination pathways.
    }
    \vspace{-10pt}
    \label{fig:vis_comp}
\end{figure*}

Additionally, we present qualitative results in \figref{fig:imagination_sequence}.
In the top row, we decode the latent states before and after the jumpy state transitions back to the pixel space.
To better understand how affordance maps facilitate the jumpy state transitions and whether they can provide effective goal-conditioned guidance, the bottom rows visualize the affordance maps reconstructed from the latent states.
These visualizations demonstrate that the proposed world model can adaptively determine when to utilize long-term imagination based on the current visual observation. Furthermore, the generated affordance maps align effectively with areas that are highly relevant to the final goal, thereby enabling the agent to perform more efficient policy exploration.

\vspace{-3pt}
\subsection{Model Analyses}
\vspace{-3pt}
% \paragraph{Comparison between series and parallel connections .}

\paragraph{Ablation studies.}
\begin{wrapfigure}{r}{0.42\textwidth}
\vspace{-15pt}
    \includegraphics[width=\linewidth]{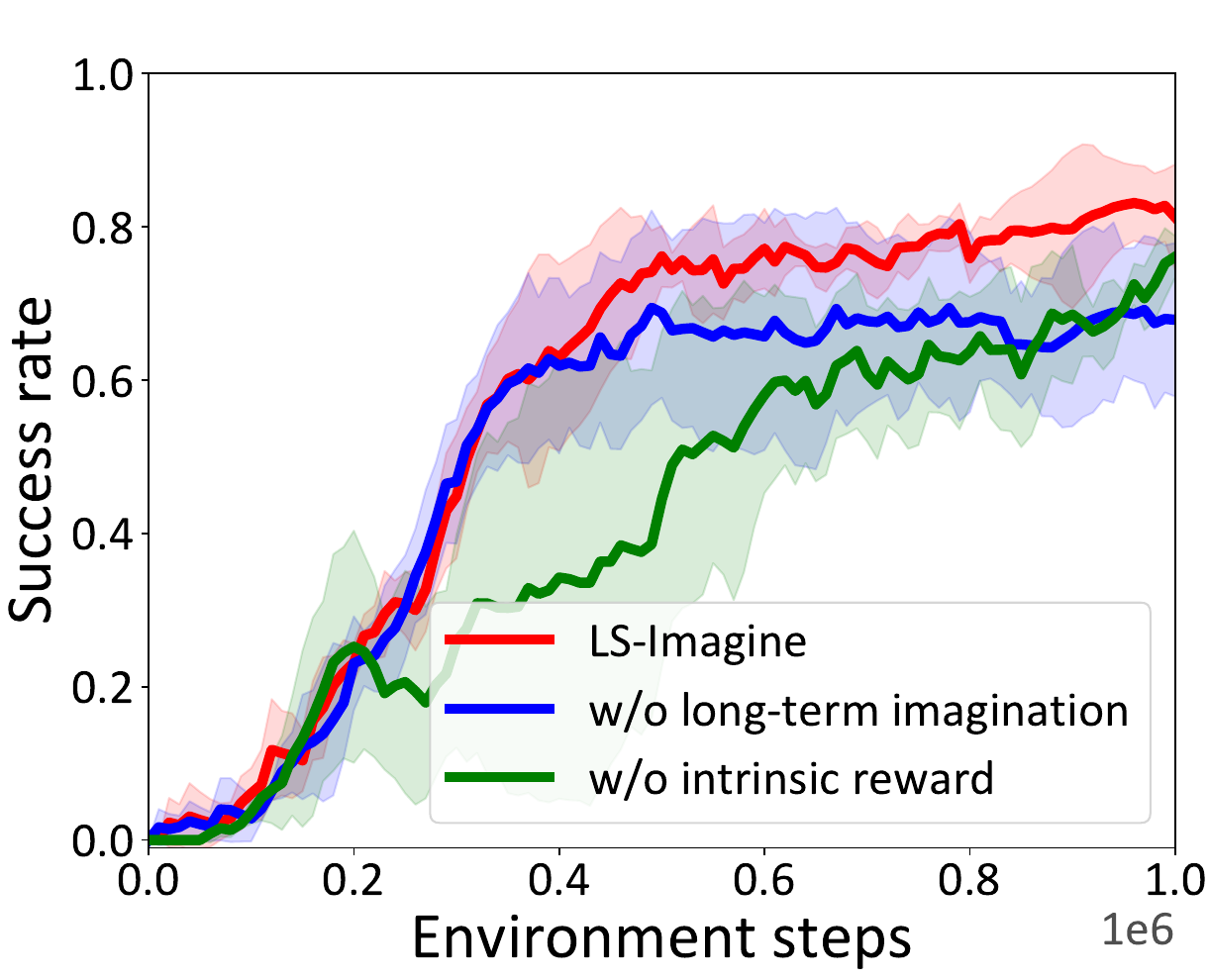}
    \vspace{-20pt}
    \caption{Ablation study results.}
    \label{fig:abl_study}
    \vspace{-5pt}
\end{wrapfigure}
We conduct the ablation studies to validate the effect of the affordance-driven intrinsic reward and long short-term imagination.
\figref{fig:abl_study}~ presents corresponding results in the challenging MineDojo tasks.  
% It can be observed that the performance of \model{} declines when we remove two main components introduced in this paper.
%
As shown by the blue curve, removing the long-term imagination of \model{} leads to a performance decline, which indicates the necessity of introducing long-term imagination and switching between it and short-term imagination adaptively. 
%Benefiting from long short-term imagination, agents can achieve higher sample efficiency during training.
%
For the model represented by the green curve, we do not employ affordance-driven intrinsic reward. It shows that the affordance-driven intrinsic reward also plays an important role during the early training stage of agents. 
Additionally, unlike the MineCLIP reward being calculated based on a series of states, the affordance-driven intrinsic reward relies solely on a single independent state. This approach enables a more accurate estimation of the reward for the post-jumpy-transition state.

% a smaller value results in a lower success rate in the later stages. On the other hand, a larger value led to lower sampling efficiency during training but resulted in a higher success rate in the end. The reason for this behavior is that setting the long-term branch loss scale too high or too low will cause the world model to overly focus on one branch during training. Therefore, in our experiments, we select $\beta_{\text{long}} = 1$ to balance the training between the two branches, aiming for good performance in both long-term and short-term imagination.
%
% For $\alpha$, a smaller value caused the affordance-driven intrinsic reward to have a lower proportion in the overall reward, which led to lower sampling efficiency during training. However, if $\alpha$ was set too high, the reward imbalance caused a decline in the overall training performance. Therefore, we set $\alpha=1$ in our experiments to balance the MineCLIP reward and the affordance-driven intrinsic reward, achieving the best training performance.
%

\vspace{-8pt}
\paragraph{Alternative pathways of mixed imaginations.}
It is worth highlighting that the long short-term imagination is implemented sequentially. 
In ~\figref{fig:series_implementation} in the appendix, we provide a visualization illustrating how the agent sequentially performs short-term and long-term imaginations within a single sequence.
Alternatively, as illustrated in \figref{fig:parallel_implementation}, we could structure long- and short-term imagination pathways in parallel. 
Specifically, we begin by applying short-term imagination within a single sequence. For each predicted state, we examine the jumping flag: If $\hat{j}_t = 1$, we initiate a new imagination sequence starting from the post-jump state, which is predicted by the long-term transition model and the dynamics predictor.
In other words, whenever a long-term state jump occurs, the world model generates a new sequence from the post-jump state, while the intermediate state transitions within the sequence are governed exclusively by short-term dynamics.
Importantly, we optimize the actor independently for each sequence, ensuring that there is no gradient or value transfer between sequences.
To evaluate the advantages of using sequential long short-term imagination, we conduct an experimental comparison between \model{}~(\textit{series}) and \model{}~(\textit{parallel}). 
\figref{fig:sequencial_and_parallel} shows that the \model{}~(\textit{series}) outperforms \model{}~(\textit{parallel}) by large margins.
This implies that the parallel imagination sequences are independent of one another, meaning that the sequence starting with a post-jumping state does not guide the prior-jumping transitions.

In the appendix, we further include (i) experiments on the long-horizon ``Tech Tree'' task, (ii) analyses of the long-term imagination frequency and corresponding state jumping intervals $\hat{\Delta}_t$ predicted by the model, and (iii) visualization of affordance maps with occluded target objects.
\vspace{-3pt}
\section{Related Work}
\vspace{-3pt}

\paragraph{Visual MBRL.}
%近些年 visual rl 用的越来越多，之前的 RL 是处理低维算法的。
Recently, learning control policies from images, \ie visual RL has been used widely, whereas previous RL algorithms learn policies from low-dimensional states. Existing approaches can be grouped by the use of model-free RL methods~\citep{laskin2020curl,schwarzer2021pretraining,stooke2021decoupling,xiao2022masked,parisi2022unsurprising,yarats2022mastering,zheng2024taco} or model-based RL methods~\citep{hafner2019learning,hafner2020dream,hafner2021mastering,seo2022reinforcement,pan2022iso,zhang2023storm,mazzaglia2023choreographer,micheli2023transformers,zhang2023predictive,ying2023reward,seo2023multi,alonso2024diffusion,hansen2024td,wang2024making}. 
The following methods specifically enhance the modeling of long-term dynamics in visual MBRL.
\cite{lee2024dreamsmooth} proposed the prediction of temporally smoothed rewards to address long-horizon sparse-reward tasks.
R2I~\citep{samsami2024mastering} improves long-term memory and long-horizon credit assignment in MBRL.
%MV-MWM~\citep{seo2023multi} employs a multi-view masked autoencoder for representation learning and subsequently trains a world model based on the representations.
%
Unlike existing methods, our work presents a long short-term world model architecture specifically designed for visual control in open-world environments.

\vspace{-8pt}
\paragraph{Affordance maps for robot learning.}
Our work is also related to the affordance map for robot learning~\citep{mo2021where2act,jiang2021synergies,yarats2021image,mo2022o2o,geng2022end,xu2022partafford,wang2022adaafford,wu2022vat,ha2022flingbot,xu2022universal,cheng2024empowering,lee2024affordance,li2024manipllm}.
Where2Explore~\citep{ning2023where2explore} introduces a cross-category few-shot affordance learning framework that leverages the similarities in geometries across different categories. 
%
%这个可以具体说下，题目中有 visual affordance，ICLR 23
DualAfford~\citep{zhao2023dualafford} learns collaborative actionable affordance for dual-gripper manipulation tasks over various 3D shapes.
%feifei 的工作具体介绍下 VoxPoser
VoxPoser~\citep{huang2023voxposer} unleashes the power of large language models and vision-language models for extracting affordances and constraints of real-world manipulation tasks, which are grounded in 3D perceptual space.
VRB~\citep{bahl2023affordances} trains a visual affordance model with videos of human interactions and deploys the model in real-world robotic tasks directly.
\cite{qi2020learning} adopted a spatial affordance map that is trained by interacting with the environment for navigation.
However, our approach distinguishes itself by employing visual observation to generate affordance maps as guidance to mitigate the low exploration efficiency in open-world environments.

\vspace{-8pt}
\paragraph{Hierarchical methods.}
Like our approach, Director~\citep{hafner2022deep} learns hierarchical behaviors in the latent space, which adopts high-level policy~(\textit{manager}) to produce latent goals to guide low-level policy~(\textit{worker}).
Dr. Strategy~\citep{hamed2024dr} proposes strategic dreaming with latent landmarks to learn a highway policy that enables the agent to move to a landmark in the dream.
\cite{gumbsch2024learning} presented a hierarchy of world models, which perform high-level and low-level prediction adaptively, and the high-level predictions depend on the low-level predictions.
Our method distinguishes itself by generating affordance maps through image zoom-in to encourage the agent to explicitly execute long-term imagination in the world model.

\vspace{-3pt}
\section{Conclusions and Limitations}
\vspace{-3pt}

In this paper, we presented a novel approach to overcoming the challenges of training visual reinforcement learning agents in high-dimensional open worlds. By extending the imagination horizon and leveraging a long short-term world model, our method facilitates efficient off-policy exploration across expansive state spaces. The incorporation of goal-conditioned jumpy state transitions and affordance maps allows agents to better grasp long-term value, enhancing their decision-making abilities. Our results demonstrate substantial improvements over existing state-of-the-art techniques in MineDojo, highlighting the potential of our approach for open-world reinforcement learning and inspiring future research in this domain.

A limitation of LS-Imagine is the computational overhead it introduces. Additionally, its effectiveness has only been validated in 3D navigation environments with embodied agents. We aim to enhance the generalization of our approach across a wider range of tasks.

\section*{Ethics Statement}
In this work, we are committed to upholding ethical research practices. This work does not involve human subjects, personal data, or sensitive information. All environments and datasets used are synthetic and publicly available. We recognize the potential for reinforcement learning models to be misused, particularly in decision-making scenarios where unintended outcomes could arise. To mitigate these risks, we emphasize responsible deployment and encourage careful consideration of the broader impact of such systems, restricting the use of our work strictly to research purposes.

\section*{Reproducibility Statement}
We prioritize the reproducibility of our work. All results can be reproduced on publicly available RL environments by following the experimental details presented in \secref{sec:experiment} and \appref{sec:hyper}. We provide the source code at \textcolor{magenta}{\url{https://github.com/qiwang067/LS-Imagine}}.

\section*{Acknowledgments}
This work was supported by the National Natural Science Foundation of China (Grants 62250062, 62302246), the Smart Grid National Science and Technology Major Project (Grant 2024ZD0801200), the Shanghai Municipal Science and Technology Major Project (Grant 2021SHZDZX0102), the Fundamental Research Funds for the Central Universities, and the CCF-Tencent Rhino-Bird Open Research Fund.
Additional support was provided by the Natural Science Foundation of Zhejiang Province, China (Grant LQ23F010008), the High Performance Computing Center at Eastern Institute of Technology, Ningbo, and Ningbo Institute of Digital Twin.

\bibliography{ref}
\bibliographystyle{iclr2025_conference}

\clearpage
\appendix

\section*{Appendix}
\section{Environment Details}
\label{sec:env_detail}
As illustrated in \tabref{tab:setting_mc}, \textit{language description} is employed for calculating the MineCLIP reward~\citep{fan2022minedojo}. \textit{Initial tools} are the items provided in the inventory at the beginning of each episode. 
\textit{Initial mobs and distance} specifies the types of mobs present at the start of each episode and their initial distance from the agent.
\textit{Max steps} refers to the maximum allowed steps per episode.

\begin{table*}[h]
\vspace{-10pt}
\caption{Details of the MineDojo tasks.} 
\label{tab:setting_mc}
\vskip 0.05in
\setlength\tabcolsep{2pt}
\begin{center}
\footnotesize
\begin{tabular}{lcccc}
\toprule
Task & Language description & Initial tools & Initial mobs and distance& Max steps  \\
\midrule
Harvest log in plains & ``Cut a tree.'' & --  &-- & 1000\\
Harvest water with bucket & ``Obtain water.'' & bucket &-- & 1000\\
Harvest sand & ``Obtain sand.'' & -- &-- & 1000\\
Shear sheep & ``Obtain wool.'' & shear &sheep, 15 & 1000\\
Mine iron ore & ``Mine iron ore.'' & stone pickaxe & -- & 2000\\
\bottomrule
\end{tabular}
\end{center}
\vspace{-10pt}
\end{table*}

\section{Compared Methods}
\label{sec:compared_methods}
We compare \model{} with strong Minecraft agents, including: 
\begin{itemize}[leftmargin=*]
    \vspace{-5pt}
    \item \textit{DreamerV3}~\citep{hafner2023dreamerv3}:
    An MBRL approach that learns directly from the step-by-step imaginations of future latent states generated by the world model. 
    % This latent representation allows agents to imagine trajectories simultaneously.
    % \vspace{-2pt}
    \item \textit{VPT}~\citep{baker2022video}: A foundation model designed for Minecraft trained through behavior cloning, on a dataset consisting of $70{,}000$ hours of game playing collected from the Internet.   
    \item \textit{STEVE-1}~\citep{lifshitz2023steve}: An instruction-following Minecraft agent that translates language instructions into specific goals. 
    % The development process begins by training a goal-conditioned policy using the contractor dataset. This policy serves as the low-level controller. Subsequently, Steve-1 employs a language-labeled dataset to translate language instructions into specific goals. 
    To evaluate its effectiveness, we assess Steve-1's zero-shot performance on our tasks by supplying it with task instructions.
    % \item \textbf{CURL}~\citep{laskin2020curl}: 
    % A model-free RL approach that leverages contrastive learning to extract high-level features from raw pixel data.
    % \vspace{-5pt}
    \reb{\item \textit{Director}~\citep{hafner2022deep}: An agent learns hierarchical behaviors by leveraging a world model to plan within its latent space.
    \item \textit{PTGM}~\citep{yuan2024pre}: An RL method that pretrains goal-based policy and adopts temporal abstractions and behavior regularization.}
\end{itemize}

\section{Model Details}
\subsection{Environmental Interaction and Data Collection}
\label{sec:env_interact}

\reb{
To train \model{}'s world model, we collect both short-term and long-term transition data through interactions with the environment. 
As shown in \figref{fig:env_interact}, at each time step $t$, the agent interacts with the environment following the current policy. 
At each time step, the data buffer collects a tuple $\mathcal{D}_t$, which includes 
$(o_t, a_t, \mathcal{M}_t, r_t, c_t, j_t, \Delta_t, G_t)$:
% $\{o_t, a_t, c_t, \mathcal{M}_t, r_t\}$
\begin{itemize}[leftmargin=*]
\vspace{-5pt}
    \item $o_t$ represents the observed image.
    \item $a_t$ represents the agent's action taken given $o_t$.  
    \item $\mathcal{M}_t$ is the affordance map generated by a multimodal U-Net given $o_t$ and task instructions $I$. 
    \item $r_t$ is defined in Sec. \ref{sec:final_reward}, which is the immediate reward computed as a weighted sum of the sparse environmental reward $r^{\text{env}}_t$ after executing $a_{t-1}$, the MineCLIP reward $r_t^{\text{MineCLIP}}$ from a pretrained scoring model~\citep{fan2022minedojo}, and the intrinsic reward $r_{t}^{\text{intr}}$ defined in \eqref{eq:intrinsic_reward} and based on $\mathcal{M}_t$.
    \item $c_t$ is the continuation flag received from the environment, which indicates whether further interaction is required after this step.
    \item $j_t$ is the jumping flag, which is used to train the world model to trigger long-term imagination during model-based behavior learning. We first estimate the jumping probability $P_{\text{jump}}$ using \eqref{eq:jump_prob} based on $\mathcal{M}_t$. To stabilize training, we establish a dynamic threshold $P_\text{thresh}$, which accounts for the varying guidance strength provided by the affordance map across different tasks, resulting in task-specific distributions of $P_\text{jump}$. Specifically, from the beginning of training, we store the $P_\text{jump}$ values for every interaction step in a dedicated buffer. The threshold $P_\text{thresh}$ is then dynamically calculated as the mean of all $P_\text{jump}$ values currently in the buffer plus their standard deviation. This dynamic adjustment ensures that the threshold adapts to the characteristics of the task and remains robust throughout training. If $P_{\text{jump}} > P_\text{thresh}$, we set $j_t=1$; otherwise, $j_t=0$.
    \item $\Delta_t$ represents the expected number of step intervals in the jumpy state transitions during long-term imaginations. Specifically, we set $\Delta_t = 1$ by default, corresponding to a short-term transition.
    \item $G_t$ represents the expected cumulative reward between the pre- and post-jump states when long-term imagination occurs. Specifically, for a short-term transition, we set $G_t = r_t$ by default.
\end{itemize}
If $j_t=0$, $\mathcal D_t$ is defined as the starting point of a \textit{short}-term transition within the pair $(\mathcal D_{t}, \mathcal D_{t+1})$. During world model training, $(\mathcal D_{t}, \mathcal D_{t+1})$ is replayed to train the related modules associated with short-term dynamics.
Once we obtain $j_t=1$ during interactions, we define the current step as the starting point of a simulated \textit{long}-term transition $(\mathcal D_t,\mathcal D^\prime_{t+1})$. 
Notably, we use $\mathcal D^\prime_{t+1}$ to differentiate from its short-term counterparts. 
%
% We have $\mathcal D_{t}=\{o_t, a_t, \mathcal{M}_t, r_{t}, c_t, j_t=1, \Delta_t, G_t\}$. Specifically, the current action $a_t^\prime$ is unrelated to the next-step long-term state jumping, but we include it here to maintain a uniform expression.

\begin{figure}[t]
\vspace{-10pt}
    \centering
    \includegraphics[width=\textwidth]{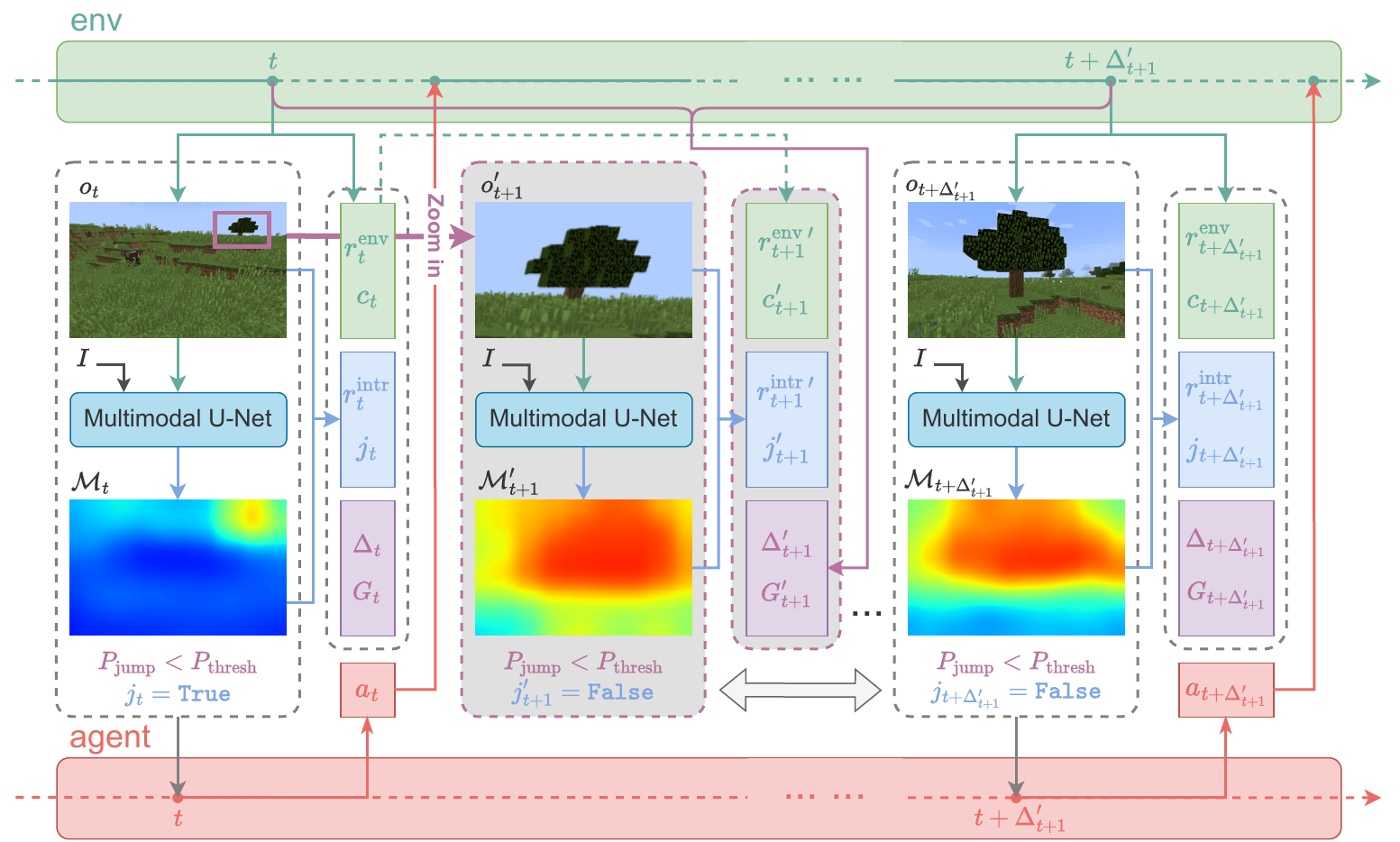}
    \vspace{-15pt}
    \caption{
    Environmental interaction and data collection. 
     }
    % The overall transfer learning pipeline of CoWorld involves two model-based RL agents and three training stages. Please refer to the text in \cref{sec:overall_pipline} for detailed descriptions.
    \label{fig:env_interact}
    \vspace{-10pt}
\end{figure}

We define $\mathcal D^\prime_{t+1}=(o_{t+1}^\prime, a_{t+1}^\prime, \mathcal{M}_{t+1}^\prime, r_{t+1}^\prime, c_{t+1}^\prime, j_{t+1}^\prime, \Delta_{t+1}^\prime, G_{t+1}^\prime)$, where
$r_{t+1}^\prime$ and $c_{t+1}^\prime$ are computed in the same manner as in short-term tuples but with $o_{t+1}^\prime$ and $\mathcal M_{t+1}^\prime$ as inputs. 
Similarly, $a_{t+1}^\prime$ and $j_{t+1}^\prime$ are also computed in the same way as in short-term tuples.
We record them in the data buffer for better training of the reward predictor and the jump predictor. 

The next question is how to annotate $\Delta_{t+1}^\prime$, $G_{t+1}^\prime$, and $o_{t+1}^\prime$ to train the long-term branch.
\begin{itemize}[leftmargin=*]
\vspace{-5pt}
    \item $o_{t+1}^\prime$ is a simulated image rather than a real-captured image. It is obtained by cropping the original observation $o_t$ based on the high-value regions in the affordance map $\mathcal{M}_{t}$.
    \item $\Delta_{t+1}^\prime$ is an estimation of the number of real interaction steps between the \textit{pre-jump} state and the \textit{post-jump} state. Since the post-jump state is not real data obtained from the environment, we first identify a real state that closely resembles the post-jump state. We then calculate the number of steps required to transition from the pre-jump state to this identified real post-jump state. Specifically, we use the intrinsic reward as a measurement. Starting from the pre-jump state, during subsequent interactions with the environment, if the agent reaches a real state where the intrinsic reward satisfies $ r_{t+\Delta_{t+1}^\prime}^\text{intr} \geq r_{t+1}^{\text{intr} \ \prime}$, we take this state as the real post-jump state and take $\Delta_{t+1}^\prime$ as the long-term jumping interval. 
    \item $G_{t+1}^\prime$ is the cumulative reward within $\Delta_{t+1}^\prime$ interaction steps, \ie $G_{t+1}^\prime = \sum_{i=1}^{\Delta_{t+1}^\prime} \gamma^{i-1} r_{t+i}.$
\end{itemize}
}

\begin{figure}[t]
\vspace{-5pt}
    \centering
    \includegraphics[width=0.95\textwidth]{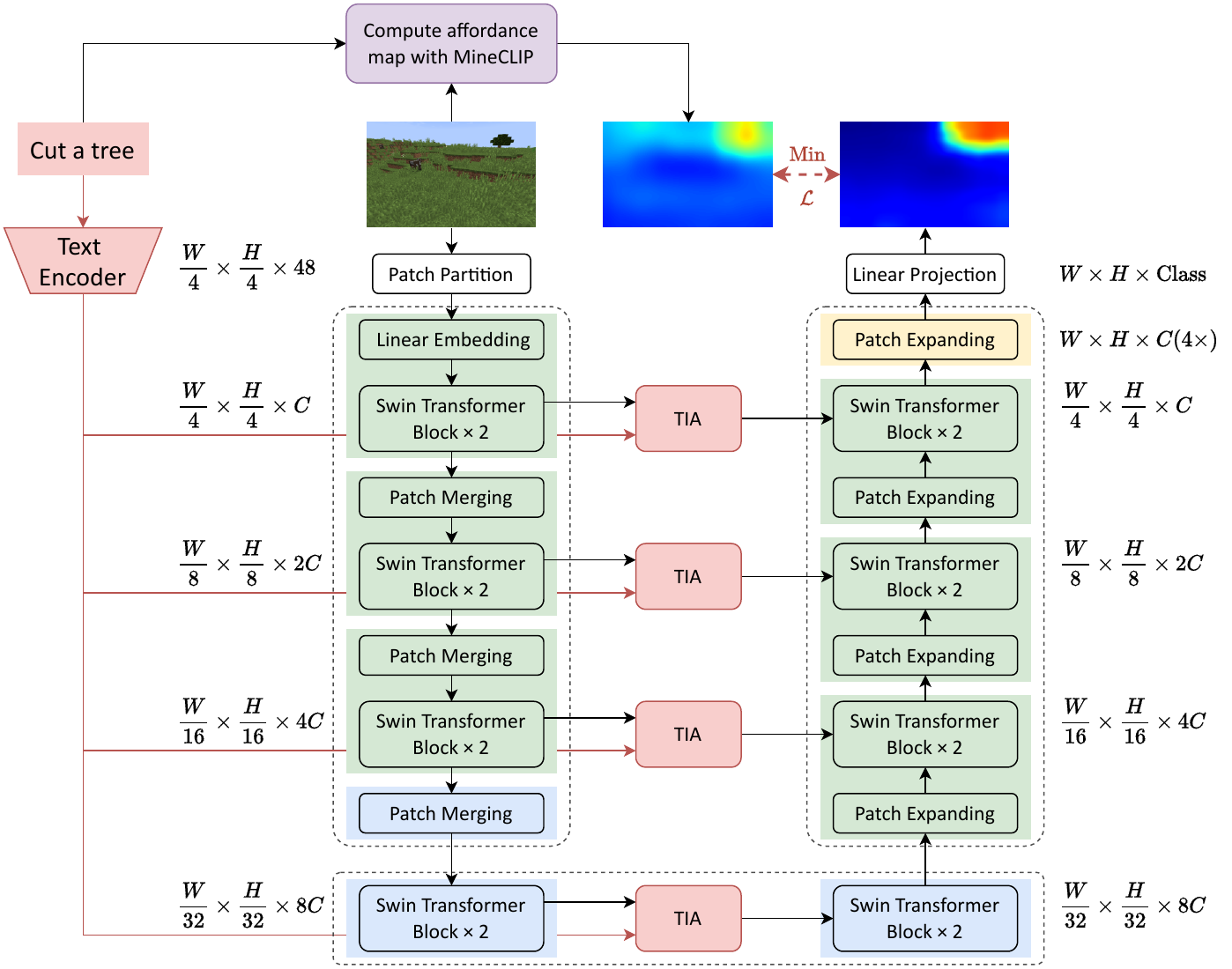}
    \vspace{-5pt}
    \caption{The architecture of multimodal U-Net.}
    % The overall transfer learning pipeline of CoWorld involves two model-based RL agents and three training stages. Please refer to the text in \cref{sec:overall_pipline} for detailed descriptions.
    \label{fig:multimodal_unet}
    \vspace{-10pt}
\end{figure}

\vspace{-3pt}
\subsection{Framework of Multimodal U-Net}
\vspace{-3pt}
As described in \secref{sec:affordance_map_gen_with_unet}, we train a multimodal U-Net to rapidly generate affordance maps based on observation images and task instructions. Our enhanced multimodal U-Net architecture, as illustrated in \figref{fig:multimodal_unet}, is based on Swin-Unet~\citep{swinunet}, a U-shaped encoder-decoder architecture built on Swin Transformer blocks. The enhanced multimodal U-Net consists of an encoder, a decoder, a bridge layer, and a text processing module. In the Swin-Unet-inspired structure, the basic unit is the Swin Transformer block. For the encoder, the input image is divided into non-overlapping patches of size $4 \times 4$ to convert the input into a sequence of patch embeddings. 
Through this method, each patch has a feature dimension of $4\times4\times3 = 48$. 
The patch embeddings are then projected through a linear embedding layer (denoted as $C$), and the transformed patch tokens are passed through several Swin Transformer blocks and patch merging layers to produce hierarchical feature representations. The patch merging layers are responsible for downsampling and increasing the dimensionality, while the Swin Transformer blocks handle feature representation learning.

For the task instruction, the text description is processed through the text encoder of MineCLIP~\citep{fan2022minedojo} to obtain text embeddings, which are integrated with the image features extracted at each layer of the encoder via the Text-Image Attention (TIA) module. The TIA module employs a multi-head attention mechanism to fuse image features (as keys and values) with text features (as queries) in a multi-scale attention-based fusion. The resulting fused text-image features are passed through the bridge layer and are subsequently combined with the corresponding features during the upsampling process in the decoder.

The decoder comprises Swin Transformer blocks and patch-expanding layers. The extracted context features are combined through the bridge layer with the multi-scale text-image features from the encoder to compensate for the spatial information lost during downsampling and to integrate the text information. Unlike the patch merging layers, the patch expanding layers are specifically designed for upsampling. They reshape the adjacent feature maps by performing a $2\times$ upsampling of the resolution, expanding the feature maps into larger ones. Finally, a patch expanding layer performs a $4\times$ upsampling to restore the resolution of the feature map to the input resolution $W\times H$ ), followed by a linear projection layer applied on the upsampled features to produce pixel-level affordance maps.

\vspace{-3pt}
\subsection{Variants of Long Short-Term Imaginations}
\begin{figure}[t]
\vspace{-5pt}
    \centering
    \subfigure[\reb{\model{}~(\textit{series})}]{\includegraphics[width=\textwidth]{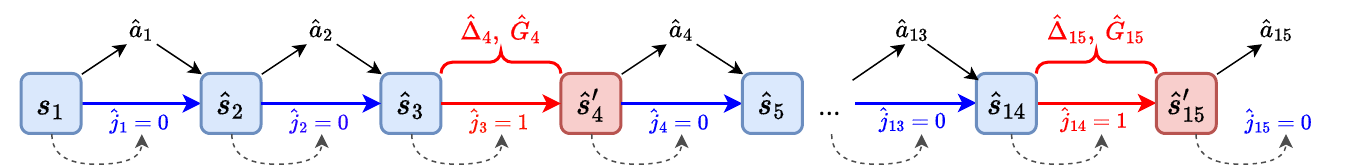} 
    \label{fig:series_implementation}
    }
    
    \subfigure[\reb{\model{}~(\textit{parallel})}]{\includegraphics[width=\textwidth]{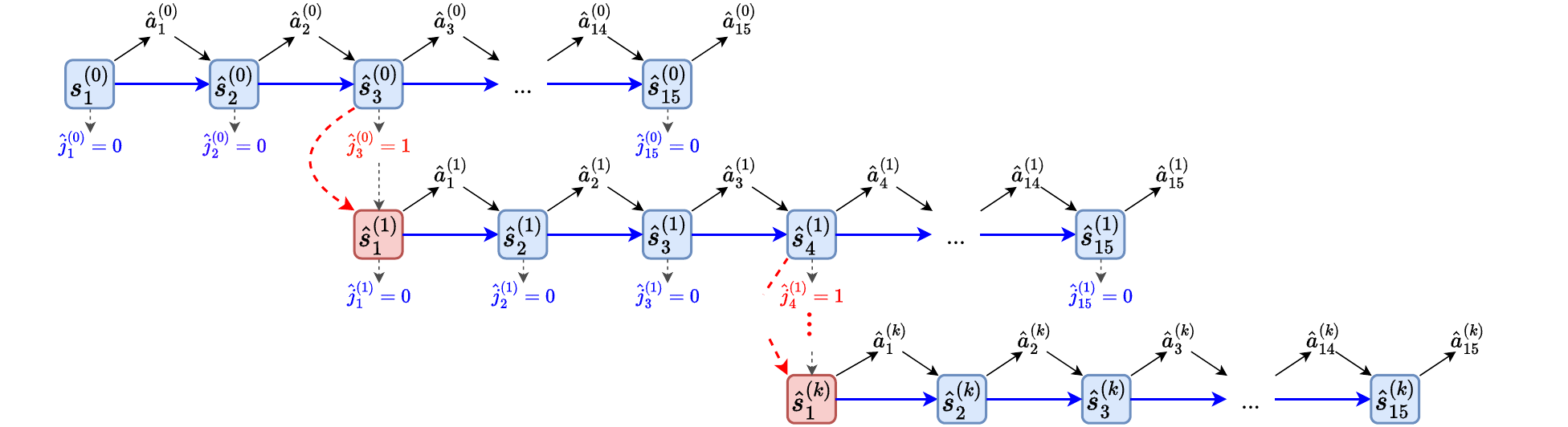}
    \label{fig:parallel_implementation}
    }
    \vspace{-15pt}
    \caption{\reb{Comparison with series and parallel variants of mixed imaginations.}
    }
    \label{fig:variant_cmp}
    \vspace{-10pt}
\end{figure}

\vspace{-3pt}
\reb{We compare two alternative pathways of the long short-term imaginations in \figref{fig:variant_cmp}.
}

\vspace{-3pt}
\subsection{Full Algorithm}
\vspace{-3pt}
\label{sec:algo}
We present the training pipeline of \model{} in Algorithm \ref{algo:overall}.

\begin{algorithm}[ht]
  \caption{The training pipeline of \model{}.}
  \label{algo:overall}
  \begin{algorithmic}[1]
  \small
  % \SetAlgoLined
  % \DontPrintSemicolon
  % \State \textbf{Hyperparameters:}{$L$: Imagination horizon; $\tau$: Sliding window size for affordance annotation}. 
  \State \textbf{Initialize} parameters $\phi,\theta, \psi$. 
\State{Compute affordance map with MineCLIP.} \Comment{Affordance map generation}
\State{Train multimodal U-Net with annotated data.}
\Comment{To enable real-time interaction with the affordance maps}
\State Train the random agent and collect a replay buffer $\mathcal{B}$.  
 \While{not converged}
    \State Sample long short-term transitions from $\mathcal{B}$.\Comment{Representation learning}  
    \State{Update the world model $\phi$ using \eqref{eq:dyn_wm_loss}.} 
    \State Generate $(s_1, \hat{a}_1, \hat{j}_1)$ using $\pi_\theta$ and $\mathcal{W}_\phi$.
    \For{time step $t=2\cdots L$} \Comment{Behavior learning}
    \If{jump flag $\hat{j}_{t-1}$} 
    \State{Generate $(s^{\prime}_t,a^{\prime}_t,c^{\prime}_t,j^{\prime}_t,\Delta^{\prime}_t,G^{\prime}_t)$ using $\pi_\theta$ and long-term imagination of $\mathcal{W}_\phi$.}
    \State{Update $(\hat{s}_t,\hat{a}_t,\hat{c}_t,\hat{j}_t,\hat{\Delta}_t,\hat{G}_t) \leftarrow (s^{\prime}_t,a^{\prime}_t,c^{\prime}_t,j^{\prime}_t,\Delta^{\prime}_t,G^{\prime}_t)$.}
    \Else
    % \State{Update $(\tilde{z}_i, \tilde{a}_i)\leftarrow (z_i, a_i)$.}
    \State{Generate $(\tilde{s}_{t}, \tilde{a}_{t},\tilde{r}_{t},\tilde{c}_{t},\tilde{j}_{t})$ using $\pi_\theta$ and short-term imagination of $\mathcal{W}_\phi$.}
    \State{Update $(\hat{s}_t,\hat{a}_t,\hat{c}_t,\hat{j}_t,\hat{\Delta}_t,\hat{G}_t)  \leftarrow (\tilde{s}_{t}, \tilde{a}_{t},\tilde{c}_{t},\tilde{j}_{t}, 1,\tilde{r}_{t})$.}
    % \State{Optimize actor $\pi_{\theta}$ using \eqref{eq:actor_loss} over the short-term imagination $\{(z_i, a_i)\}_{i=t}^{t+L}$}. 
    % \State{Optimize critic $v_{\psi}$ using \eqref{eq:critic_loss} over the short-term imagination $\{(z_i, a_i)\}_{i=t}^{t+L}$}. 
    \EndIf
    \EndFor
    \State{Calculate value estimate using \eqref{eq:value_compute}.}
    \State{Optimize actor $\pi_{\theta}$ using \eqref{eq:actor_loss} over $\{(\hat{s}_{t},\hat{a}_{t},\hat{c}_{t},\hat{j}_{t},\hat{\Delta}_{t},\hat{G}_{t})\}_{t=1}^{L}$}.
    \State{Optimize critic $v_{\psi}$ using \eqref{eq:critic_loss} over  $\{(\hat{s}_{t},\hat{a}_{t},\hat{c}_{t},\hat{j}_{t},\hat{\Delta}_{t},\hat{G}_{t})\}_{t=1}^{L}$}.
    \For{time step $t=1 \cdots T$} \Comment{Environment interaction}
    \State{Sample $\hat{a}_t \sim \pi_{\theta} \left( \hat{a}_t \mid \hat{s}_t \right)$}
    \State{$r^\text{env}_t,o_{t+1},c_t \leftarrow$ \texttt{env.step}($\hat{a}_t$)}
    \State{Generate affordance map $\mathcal{M}_t$ with multimodal U-Net for each $o_t$.} 
    \State{Calculate intrinsic reward $r^\text{intr}_t$ and jump flag $j_t$ based on the affordance map.}
    \State{Collect short-term data $(o_t,a_t,\mathcal{M}_t,r_t,c_t,j_t,\Delta_t, G_t)$.}
    \If{jumpy flag $j_t$}
    \State{Construct long-term data $(o_{t+1}^{\prime},a_{t+1}^{\prime}, \mathcal{M}^{\prime}_{t+1}, r_{t+1}^{\prime},c_{t+1}^{\prime},j_{t+1}^{\prime},\Delta_{t+1}^{\prime}, G_{t+1}^{\prime})$.}
    \EndIf
    \EndFor
    \State{Append long short-term transitions to $\mathcal{B}$.} 
    \EndWhile
\end{algorithmic}
\end{algorithm}

\vspace{-3pt}
\subsection{Clarification on Stochastic Long-Term Imagination}
\vspace{-3pt}
\reb{
One might argue that long-term imagination could skip essential intermediate steps that gradually lead to the objective, potentially resulting in a lack of learning for these crucial actions.
To address this issue, we adopt a probabilistic mechanism. Specifically, even when $
\hat{j}_t = \texttt{True}$, indicating that a long-term transition is to be executed, we implement a probability of $0.7$ for executing the jump and $0.3$ for not jumping. 
This allocation ensures a $30\%$ chance that the transition will execute the short-term imagination with gradient feedback attached to the actions. This stochastic decision-making is based on a uniform distribution, providing a balanced approach between leveraging long-term imagination and capturing essential short-term behaviors.
}

\vspace{-3pt}
\subsection{Additional Limitation}
\vspace{-3pt}

 It is worth mentioning that \model{} simulates the agent's state when approaching a target object in 3D navigation environments with embodied agents by zooming in on the observed image, and sets intrinsic rewards based on whether the agent is close to and has positioned the target object at the center of the observation. Therefore, \model{} is not suitable for environments with fixed viewpoints, 2D environments, or those where the reward mechanism is more complex than approaching objects (\eg driving).

% \reb{
% \subsection{Comparison with Hierarchical Methods}
% Our approach employs a hierarchical structure in pixel-based MBRL. 
% %
% Director~\citep{hafner2022deep} learns hierarchical behaviors in the latent space, which adopts high-level policy~(manager) to designate latent goals and achieve the goals through low-level policy~(worker).
% %
% Dr. Strategy~\citep{hamed2024dr} proposed strategic dreaming with latent landmarks to learn a highway policy that enables the agent to move to a landmark in the dream.
% %
% \cite{gumbsch2024learning} presented a hierarchy of world models, which perform high-level and low-level prediction adaptively, and the high-level predictions depend on the low-level predictions.
% %
% Our method distinguishes itself by generating affordance maps through image zoom-in to encourage the agent to execute long-term imagination in the world model.
% }

\vspace{-3pt}
\section{Additional Results}
\vspace{-3pt}

\subsection{Numerical Comparisons}
\vspace{-3pt}
\reb{
\tabref{tab:minedojo_cmp_results} compares existing approaches on the challenging MineDojo environment.
}

\begin{table*}[t]
\vspace{-5pt}
\caption{\reb{The success rate and the number of steps per episode for task completion.} 
}
\vspace{-5pt}
\label{tab:minedojo_cmp_results}
\setlength{\tabcolsep}{1mm}

\footnotesize
% \scriptsize
\begin{center}
\renewcommand\arraystretch{1.2}
%\begin{small}
% \begin{sc}
\resizebox{\textwidth}{!}{
\begin{tabular}{c|cc|cc|cc|cc|cc}
\toprule
% Model
\multirow{2}{*}{\textbf{Model}} 
& \multicolumn{2}{c|}{\textbf{Harvest log in plains}} 
& \multicolumn{2}{c|}{\textbf{Harvest water with bucket}} 
& \multicolumn{2}{c|}{\textbf{Harvest sand}} 
& \multicolumn{2}{c|}{\textbf{Shear sheep}} 
& \multicolumn{2}{c}{\textbf{Mine iron ore}} 

\\
\cline{2-11}
% & succ. (\%) & succ. step & succ. rate(\%) & first succ. step & succ. rate(\%) & first succ. step & succ. rate(\%) & first succ. step & succ. rate(\%) & first succ. step \\
& succ. (\%) & succ. step & succ. (\%) & succ. step & succ. (\%) & succ. step & succ. (\%) & succ. step & succ. (\%) & succ. step \\
\hline
% \midrule
\textbf{VPT} & 6.97 & 963.32 & 0.61 & 987.65 & 12.99 & 880.54 & 1.94 & 987.49 & 0.00 & --- \\
\textbf{STEVE-1} & 57.00 & 752.47 & 6.00 & 989.07 & 37.00 & 770.40 & 3.00 & 992.36 & 0.00 & --- \\
\textbf{PTGM} & 41.86 & 811.19 & 2.78 & 977.78 & 17.71 & 833.64 & 21.54 & 887.03 & 15.14 & \textbf{1586.03} \\
\textbf{Director} & 8.67 & 968.09 & 20.90 & 931.74 & 36.36 & 825.35 & 1.27 & 995.99 & 7.82 & 1906.31 \\
\textbf{DreamerV3} & 53.33 & 711.22 & 55.72 & 628.79 & 59.88 & \textbf{548.76} & 25.13 & 841.14 & 16.79 & 1789.06 \\
\textbf{\model{}} & \textbf{80.63} & \textbf{503.35} & \textbf{77.31} & \textbf{502.61} & \textbf{62.68} & 601.18 & \textbf{54.28} & \textbf{633.78} & \textbf{20.28} & 1748.55 \\
\bottomrule
\end{tabular}
}
% \end{sc}
%\end{small}
\end{center}
\vspace{-10pt}
\end{table*}

\vspace{-3pt}
\subsection{Analyses on Long-Term Imaginations}
\vspace{-3pt}

\begin{figure*}[t]
% \vspace{-5pt}
    \centering
    \subfigure[\reb{Jumping frequency}]{\includegraphics[height=0.22\textwidth]{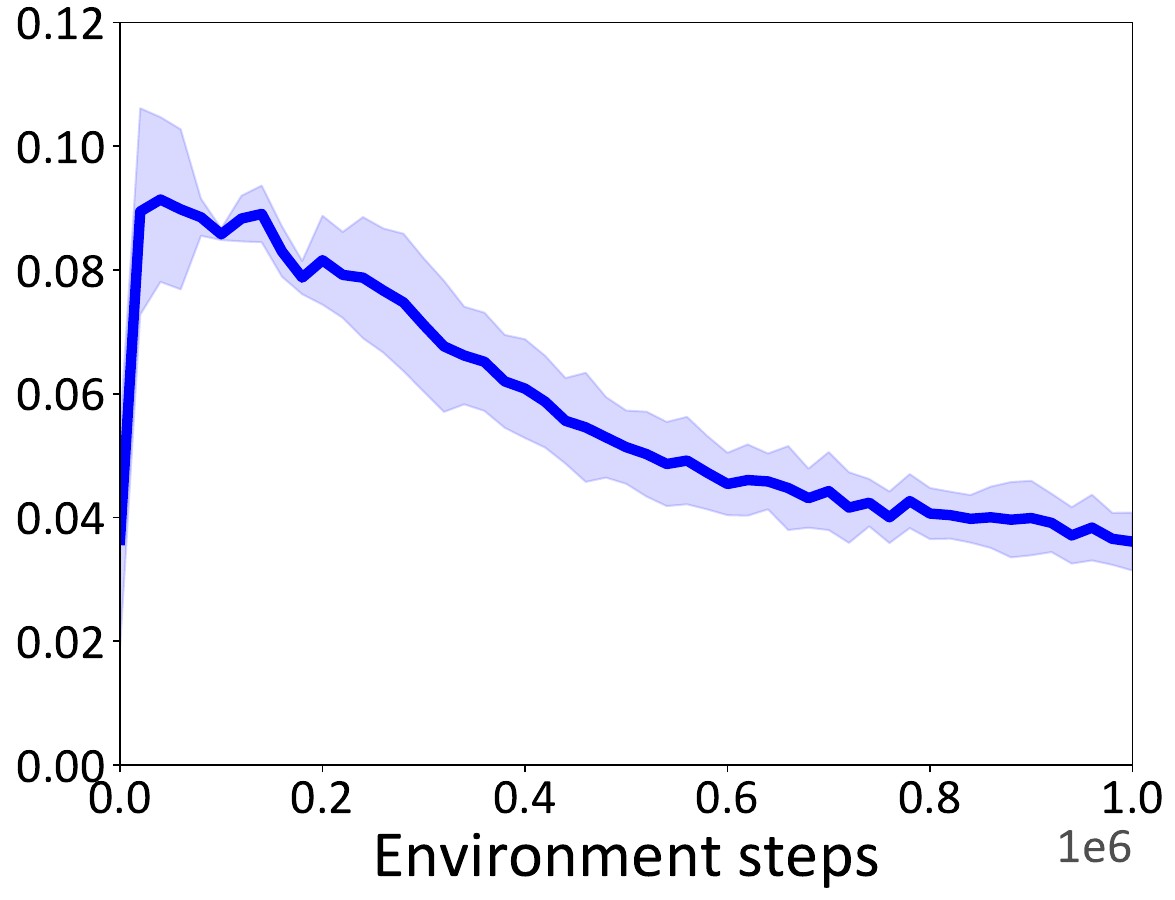}
    \label{fig:jumping_proportion}
    }
    \subfigure[\reb{Interval $\hat{\Delta}_{t}$ }]{\includegraphics[height=0.22\textwidth]{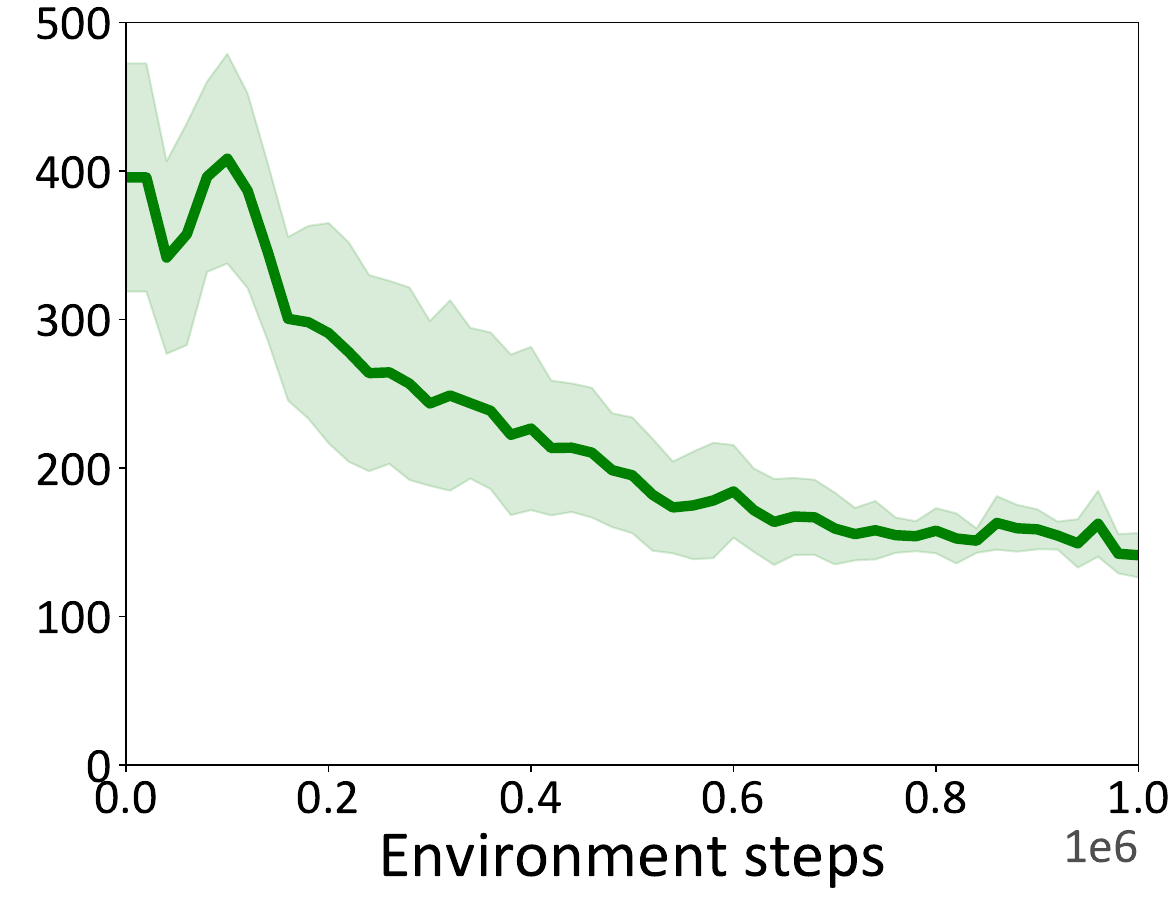}
        \label{fig:jumping_steps}
    }
    \subfigure[\reb{Dynamic threshold $P_\text{thresh}$ }]{\includegraphics[height=0.22\textwidth]{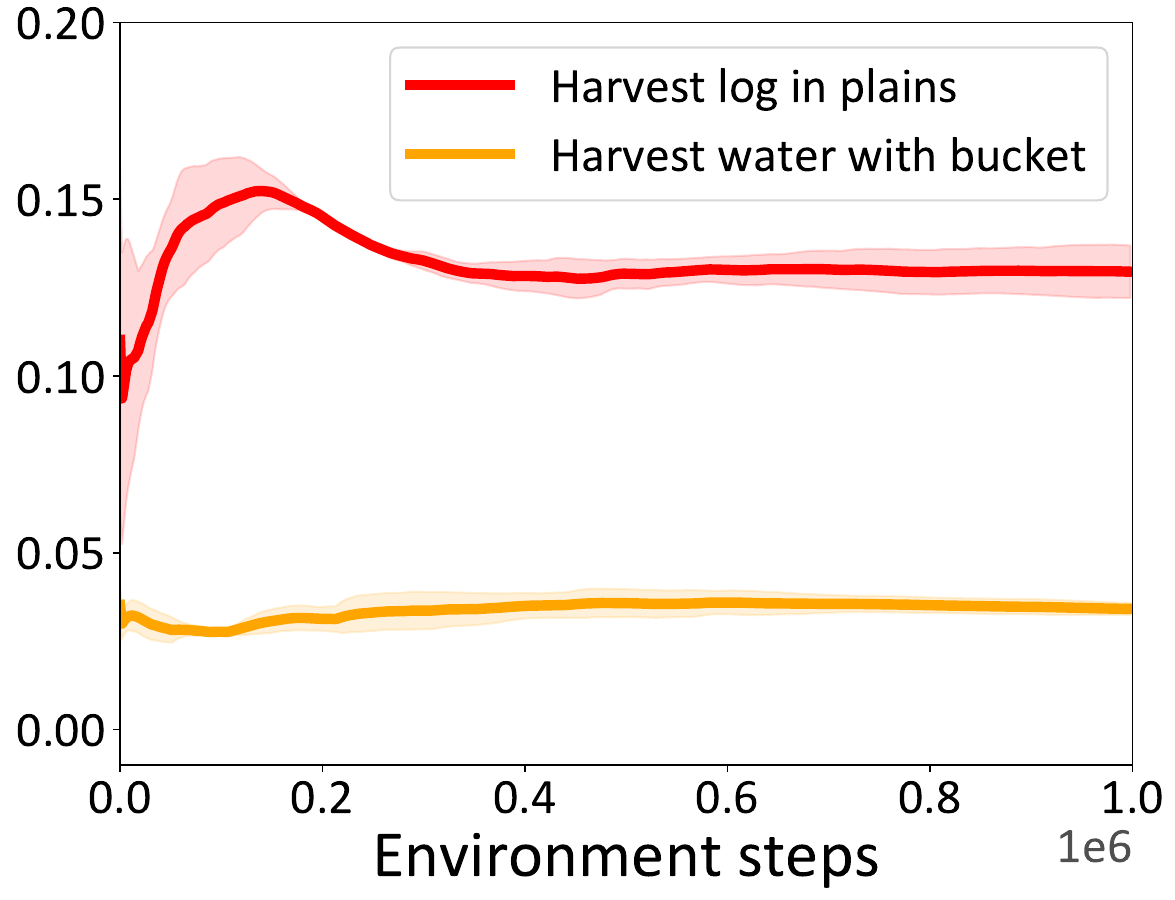}
        \label{fig:dynamic_threshold}
    }
    \vspace{-5pt}
    \caption{\reb{Analyses of long-term imaginations throughout training.}}
    \vspace{-10pt}
    \label{fig:jumping_proportion_and_steps}
\end{figure*}

\reb{
We use the task \textit{harvest log in plains} as an example to facilitate the understanding of the long short-term imagination process. 
In Figure \ref{fig:jumping_proportion}, we first track the frequency of long-term imaginations and the corresponding predicted state intervals $\hat{\Delta}_t$ throughout the training process. 
The curve shows the proportion of imagination sequences involving jumpy state transitions relative to the total number of imagination sequences.
Initially, the jumping frequency is low because the world model has not yet learned to identify when a jump is necessary based on the state. As the model’s predictions improve in the early stages of training, the frequency increases, likely due to the agent's underdeveloped policies, which result in more observations far from the goal and necessitate long-term exploration. 
Over time, as the agent learns policies that bring it closer to the target, the frequency of observations far from the goal decreases, reducing the need for jumps.

Additionally, we find that among all sequences with jumpy state transitions, the average number of jumpy transitions per sequence, within a horizon of $15$ steps, $1.02$. This indicates that, in most cases of these tasks, a single jumpy transition is sufficient to bring the agent close to the target.

In \figref{fig:jumping_steps}, we track the variations of the jumping state intervals, $\hat{\Delta}_t$, throughout training. 
At the beginning, $\hat{\Delta}_t$ is high, indicating that the policy requires many steps to reach the target. As the policy improves, fewer steps are needed to approach the target, leading to a gradual decrease in $\hat{\Delta}_t$. 
%
% This trend also reflects that \eqref{eq:value_compute}, which relies on $\hat{\Delta}_t$ for return evaluation, is fully aligned with the current policy, and there is no mismatch between the long-term transitions and the learning policy. 
Notably, as $\hat{\Delta}_t$ evolves with the updated policy, it also ensures minimal misalignment in \eqref{eq:value_compute} between the future cumulative rewards computed with jumpy imaginations and the behavior policy.

\begin{figure*}[t]
\vspace{-5pt}
    \centering
    {\includegraphics[width=\textwidth]{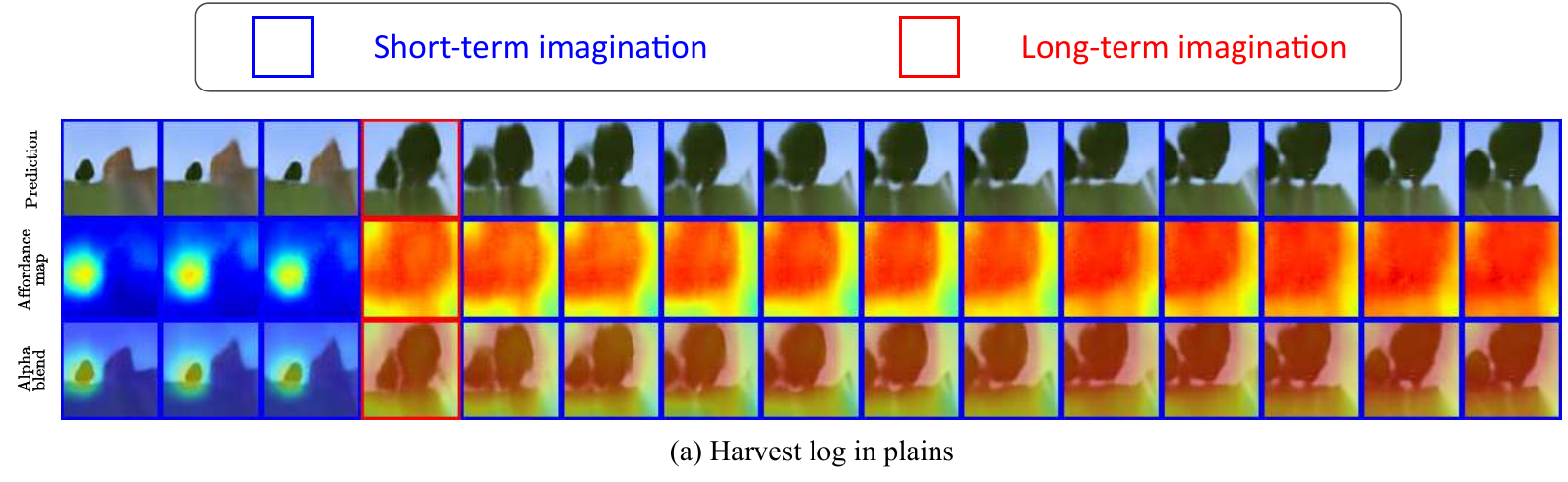}
    \vspace{-5pt}
    }
    {\includegraphics[width=\textwidth]{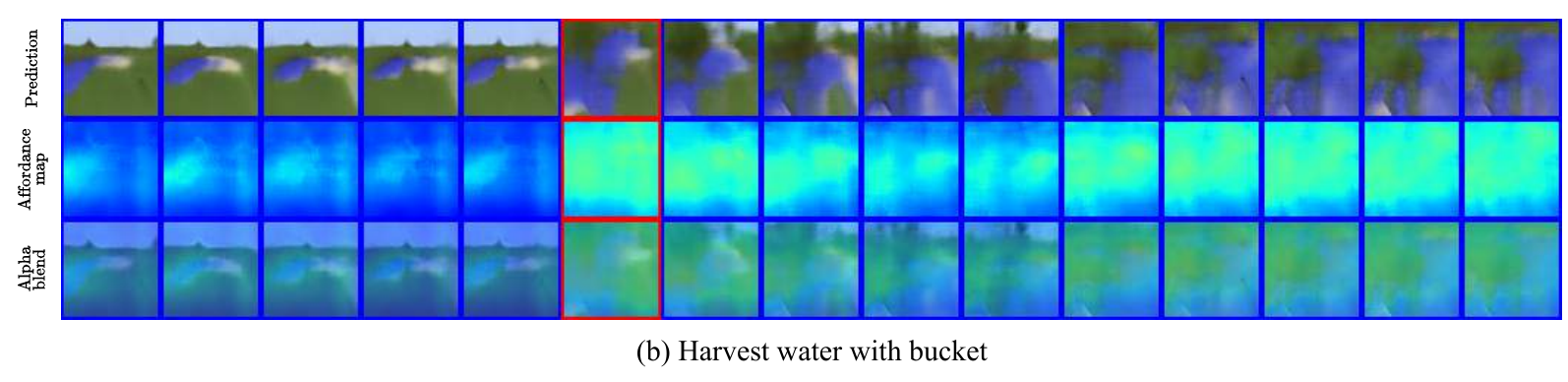}
    \vspace{-5pt}
    }
    {\includegraphics[width=\textwidth]{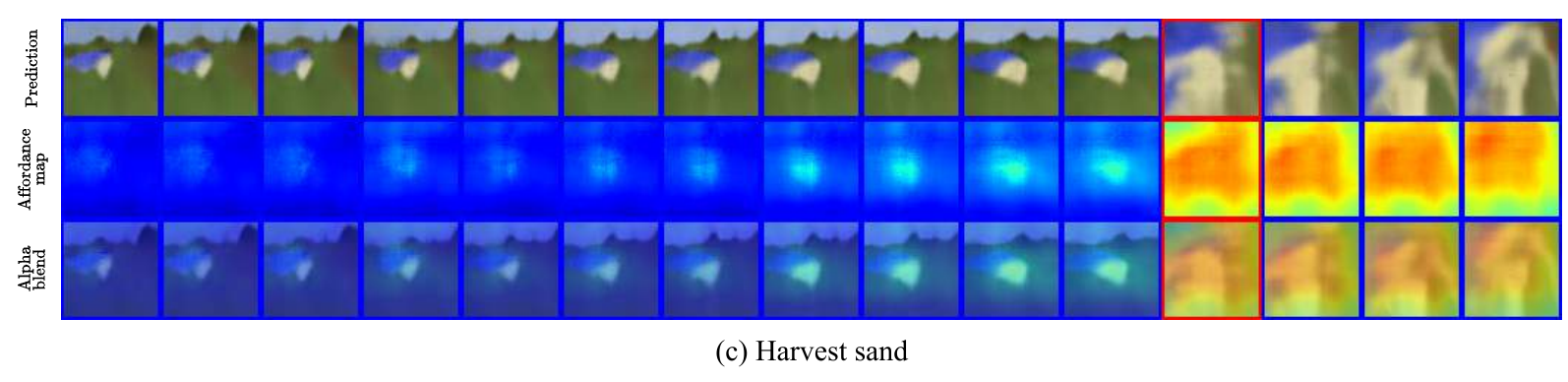}
    \vspace{-5pt}
    }
    {\includegraphics[width=\textwidth]{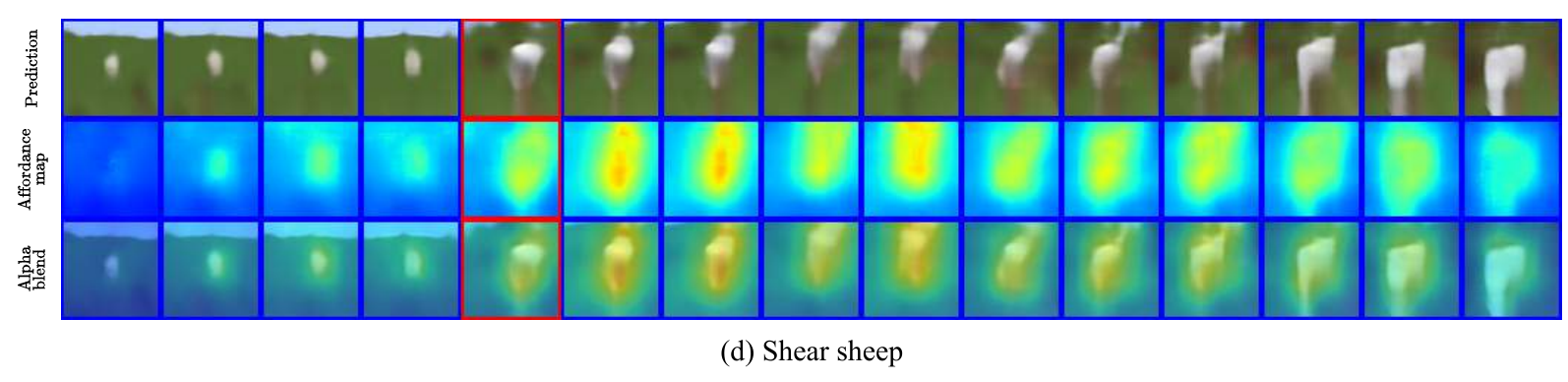}
    \vspace{-5pt}
    }
    {\includegraphics[width=\textwidth]{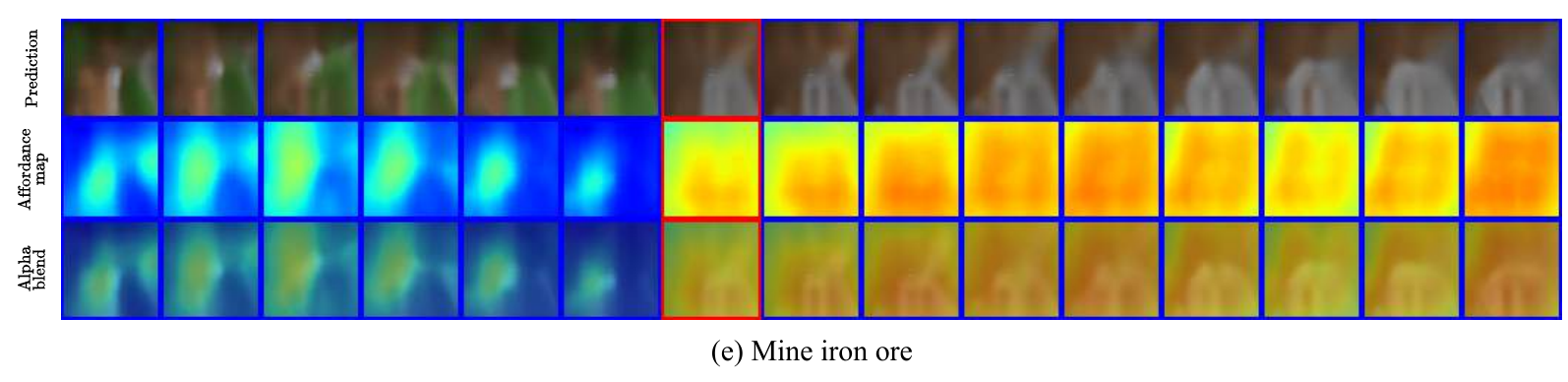}
    \vspace{-5pt}
    }
    \vspace{-15pt}
    \caption{Visualization of the complete long short-term imagination sequences.}
    \vspace{-10pt}
    \label{fig:vis_seq}
\end{figure*}

Furthermore, in \figref{fig:dynamic_threshold}, we track the variation curves of the dynamic threshold $P_\text{thresh}$ during training in different tasks, and observe that:
\begin{itemize}[leftmargin=*]
\vspace{-5pt}
    \item For task such as \textit{harvest log in plains}, the variance of $P_\text{thresh}$ is high during the early stages of training. Since $P_\text{thresh}$ serves as a temporal smoothing of $P_\text{jump}$, this reflects the significant fluctuations of $P_\text{jump}$ at the beginning of training, highlighting the importance of adopting a dynamic threshold.
    \item Across various tasks, $P_\text{thresh}$ consistently converges in the later stages of training, demonstrating its effectiveness in improving the stability of exploratory imaginations.
    \item The converged values of $P_\text{thresh}$ differ across tasks, indicating that involving an automated computation of $P_\text{thresh}$ enables us to avoid tedious hyperparameter tuning.
\end{itemize}
% \yb{todo: discuss the threshold}
}

\vspace{-3pt}
\subsection{Visualization of long short-term imaginations}
\vspace{-3pt}
% In our experiments, we not only tracked the success rate during the training process but also recorded the number of steps required to achieve success in each episode, as shown in Figure 9. The results demonstrate that compared to other baseline methods, our \model{} consistently achieves success with significantly fewer steps. 
% This highlights the improvements in exploration efficiency and interaction effectiveness brought by the affordance map and the long short-term imagination mechanism. These components enable the agent to discover, approach, and interact with the target more rapidly.

As illustrated in \figref{fig:vis_seq}, we visualize the complete long short-term imagination sequences for the agent across various tasks. This visualization further demonstrates how the affordance map accurately identifies regions of high exploration potential in the image, and how the long short-term imagination approach provides reasonable and applicable guidance for the agent's task execution. These qualitative results reinforce the effectiveness of our method in guiding the agent toward its goal with greater precision and efficiency.

\vspace{-3pt}
\subsection{Dependence on The Visibility of Objects}
\vspace{-3pt}

\begin{figure*}[t]
% \vspace{-5pt}
    \centering
    \subfigure[Explore for a village]{\includegraphics[height=0.3\textwidth]{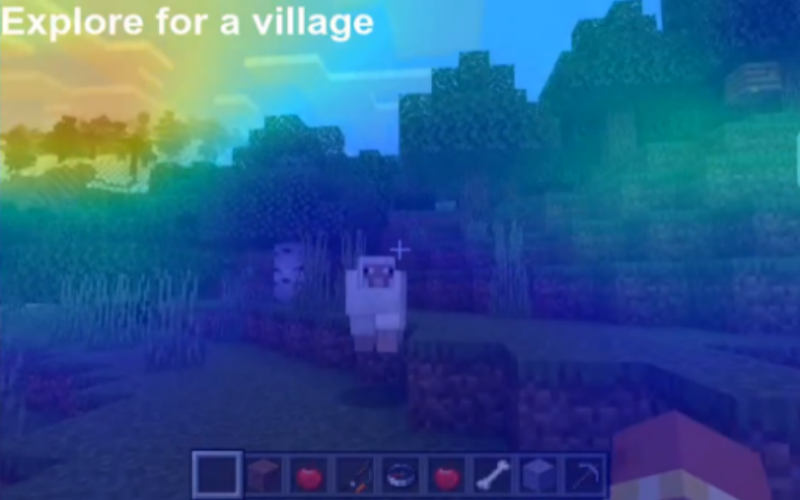}
    \label{fig:occlude_village}
    }
    \subfigure[Mine ore]{\includegraphics[height=0.3\textwidth]{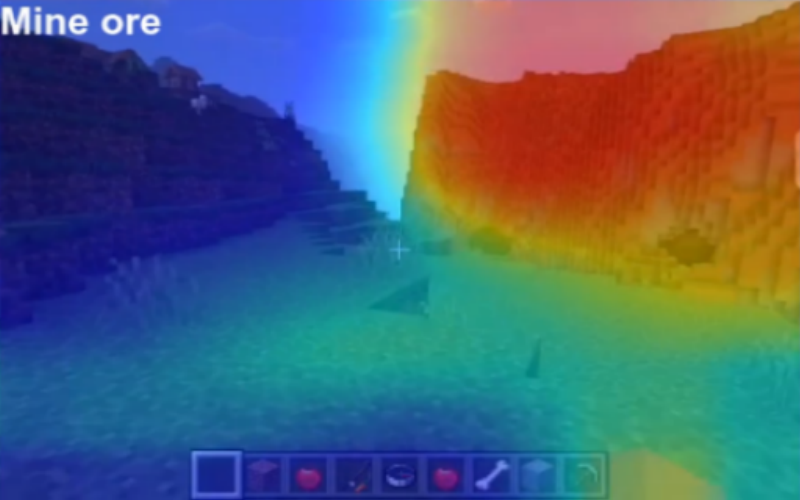}
        \label{fig:occlude_mine_ore}
    }
    \vspace{-5pt}
    \caption{\reb{Affordance maps when the target is invisible or occluded.
    }}
    % \vspace{-5pt}
    \label{fig:occlude}
\end{figure*}

\begin{figure}[t]
% \vspace{-5pt}
    \centering
    \subfigure[\reb{Success rate}]{\includegraphics[height=0.36\textwidth]{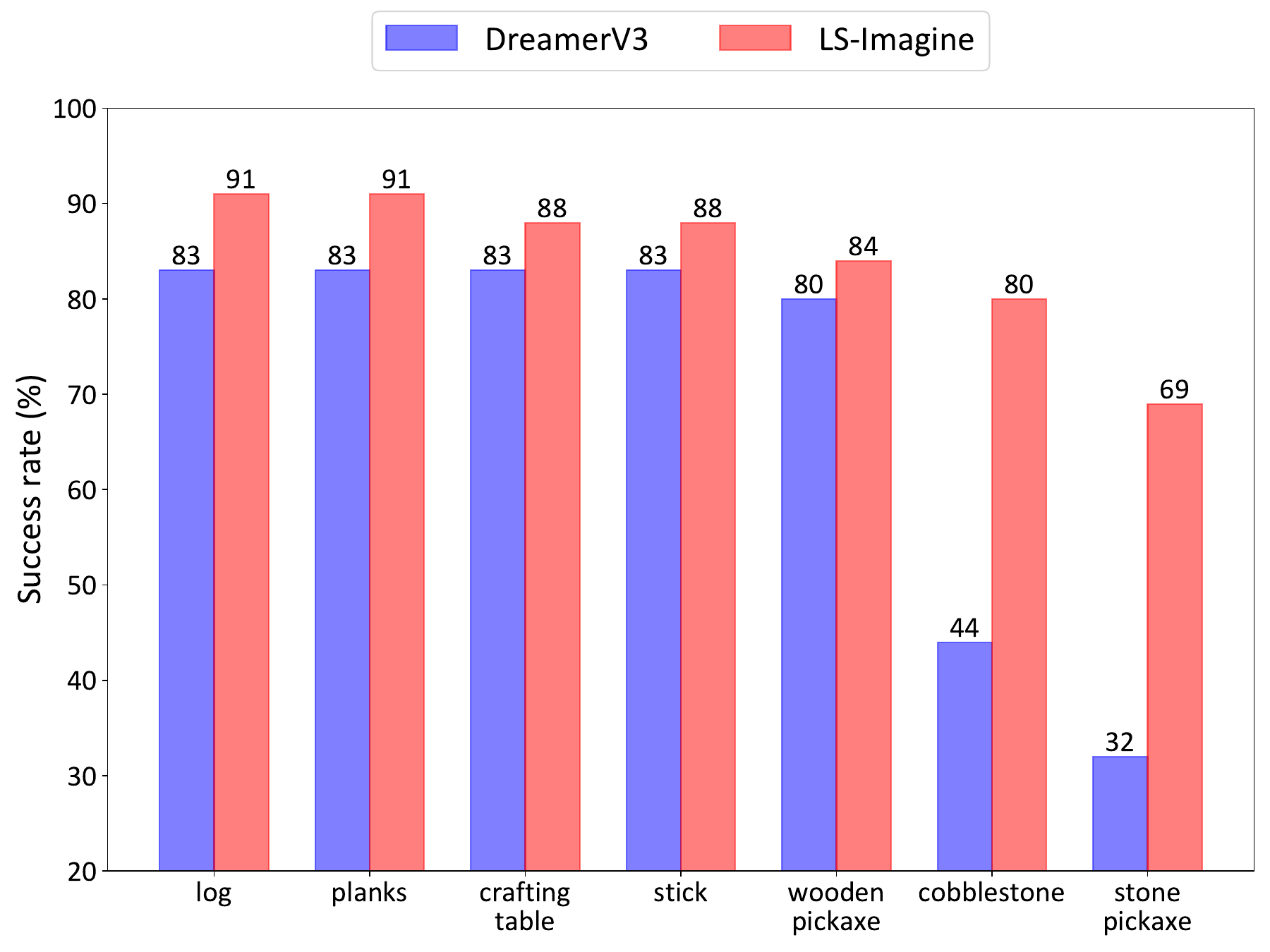}
    \label{fig:tech_tree_suc}
    }
    \subfigure[\reb{The number of steps required for task completion}]{\includegraphics[height=0.36\textwidth]{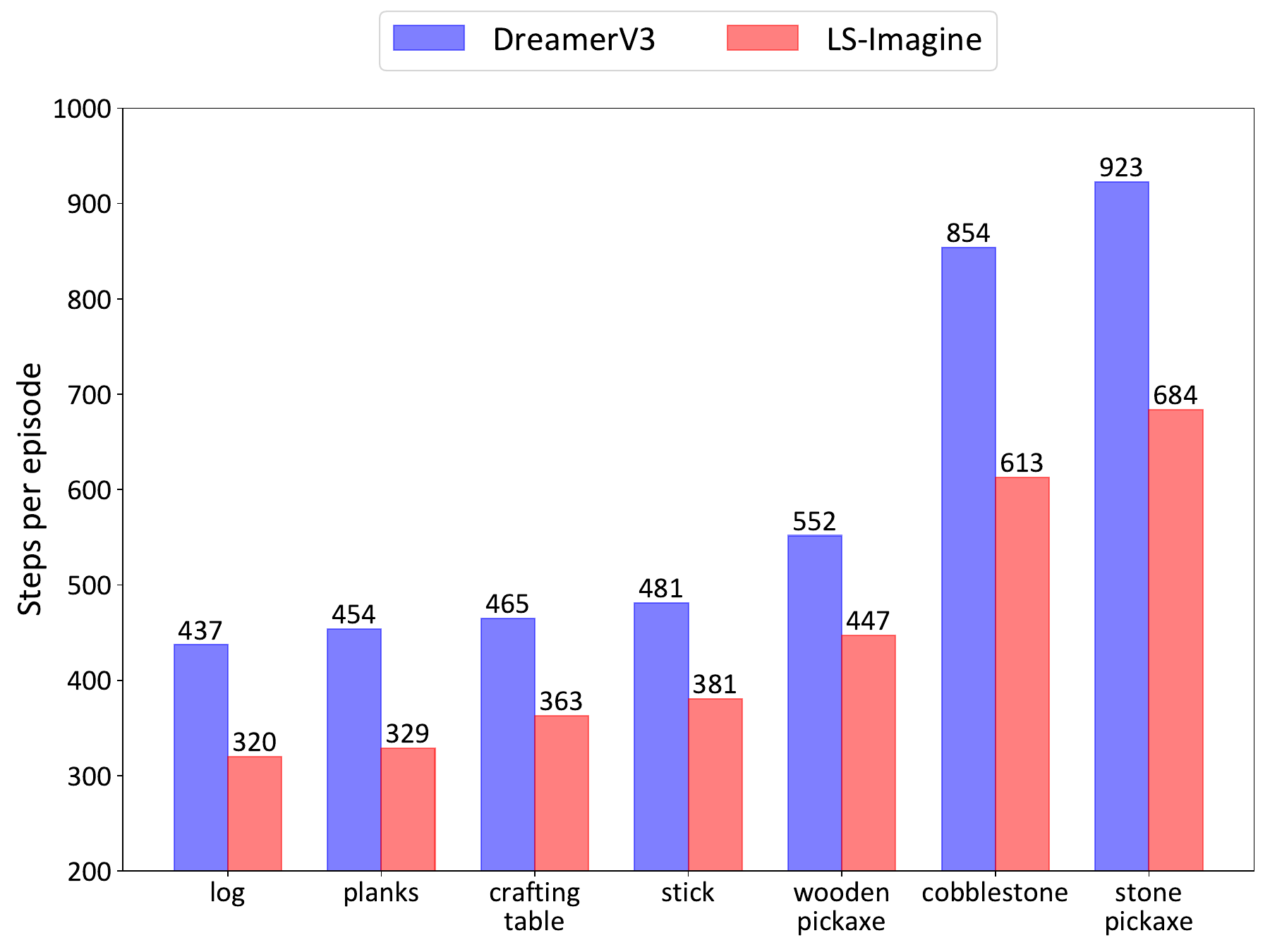}
        \label{fig:tech_tree_step}
    }
    \vspace{-8pt}
    \caption{\reb{Comparison of \model{} and DreamerV3 on a long-horizon ``Tech Tree'' task.}}
    \label{fig:tech_tree_results}
    \vspace{-5pt}
\end{figure}

% \wq{mineclip 是用 16 帧，也是提供不可见→可见的过程的先验}
% \wq{训练 Affordance map 的 UNet 的数据里面有 expert demo}

\reb{
The long-term transitions of our approach rely on the affordance map to identify high-value exploration areas. 
However, it is crucial to note that our affordance map generation method is not merely an object recognition algorithm that highlights areas only when the target is present. 
Thanks to MineCLIP's pretraining on extensive expert demonstration videos, our approach can generate affordance maps that provide guidance even when the target is completely occluded.

For instance, as illustrated in \figref{fig:occlude}, throughout the task of locating a village, the affordance map consistently provides effective guidance to the agent, suggesting exploration of the forest to the right or the open area on the left hillside, even when the village is not visible in the current observation. 
Similarly, in mining tasks where ores are typically underground, the affordance map directs the agent to dig into the mountain area on the right. 
As we can see, even when the target is occluded, the affordance map enables the agent to continue exploring effectively.
}

\vspace{-8pt}
\reb{
\paragraph{Further research direction.}
Due to the complexity of open-world environments, the affordance map may fail to provide effective guidance in scenarios that the MineCLIP model has not encountered before.
To address this issue, we plan to progressively finetune the MineCLIP model with the collected new data and introduce a new prompt to the agent: ``\textit{Explore the widest possible area to find \{target\}}'' when the affordance map fails to identify high-value areas. 
This prompt, combined with intrinsic rewards generated by MineCLIP, encourages the agent to conduct extensive exploration.
% Looking ahead, we plan to further optimize the affordance map generation algorithm. This could involve incorporating more task-specific expert demonstration data and shifting from generating value maps based solely on single-frame images to leveraging temporal sequence information. By integrating exploration history, we aim to create value maps with a deeper understanding of the environment, offering more precise guidance to support LS-Imagine in performing long-term imagination across broader environmental conditions.
}

% \begin{figure}[t]
% \vspace{-5pt}
%     \centering
%     \includegraphics[width=0.5\textwidth]{fig/mineclip_reward_cmp.pdf}
%     \vspace{-10pt}
%     \caption{\reb{MineCLIP comparison of \model{} against existing approaches in MineDojo.}
%     }
%     % The overall transfer learning pipeline of CoWorld involves two model-based RL agents and three training stages. Please refer to the text in \cref{sec:overall_pipline} for detailed descriptions.
%     \label{fig:mineclip_reward_comparion}
%     \vspace{-5pt}
% \end{figure}

\vspace{-3pt}
\subsection{Results on Long-Horizon Tasks}
\vspace{-3pt}

\reb{
To demonstrate the potential application of \model{} in more complex tasks, we conduct experiments on a ``Tech Tree'' task in MineDojo, specifically \textit{crafting a stone pickaxe from scratch}. This task involves seven subgoals: \textit{log, planks, crafting table, stick, wooden pickaxe, cobblestone, and stone pickaxe}. 
Since \model{} is primarily designed to focus on environmental interactions and task execution under fixed objectives, rather than task decomposition and planning, we adopt the DECKARD method~\citep{nottingham2023embodied} for task planning. This method provides top-level guidance, with \model{} executing the corresponding subtasks.
Each subtask was trained for $1$ million steps and then tested within $1{,}000$ steps per episode. 
The results are shown in Figure \ref{fig:tech_tree_results}, which demonstrate that our \model{} consistently outperforms DreamerV3, achieving higher success rates and requiring fewer steps to complete each subgoal.
}

\vspace{-3pt}
\subsection{Hyperparameter Analyses}
\label{sec:hyper}
\vspace{-3pt}

\begin{figure}[t]
% \vspace{-5pt}
    \centering
    \includegraphics[width=1.0\textwidth]{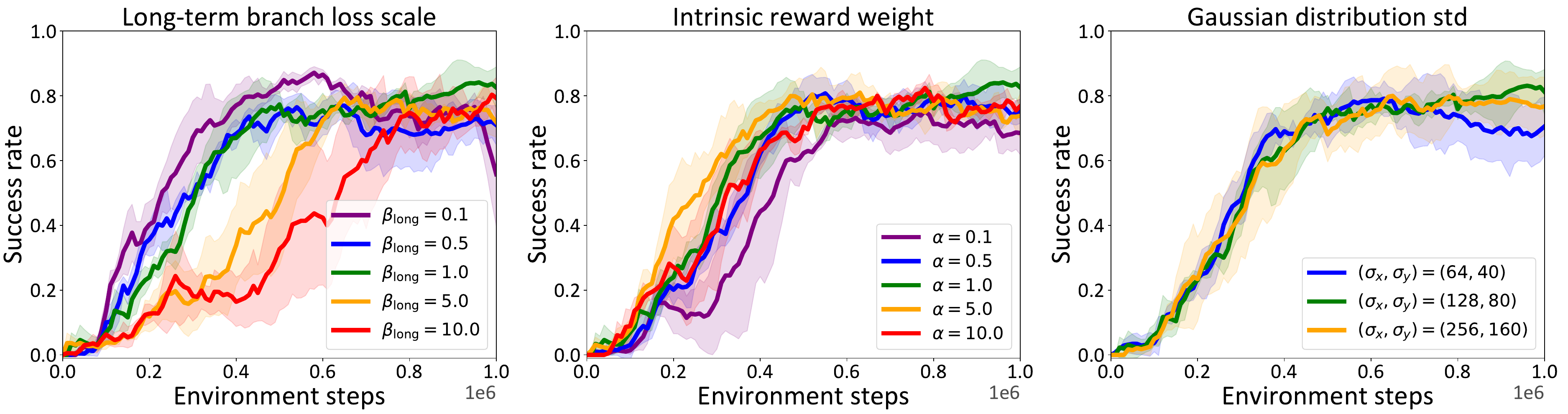}
    \vspace{-15pt}
    \caption{
    \reb{Results of hyperparameter sensitivity analyses.}
     }
    \label{fig:sensitivity_res}
    % \vspace{-5pt}
\end{figure}

\begin{figure}[t]
    \centering
    \includegraphics[width=1.0\textwidth]{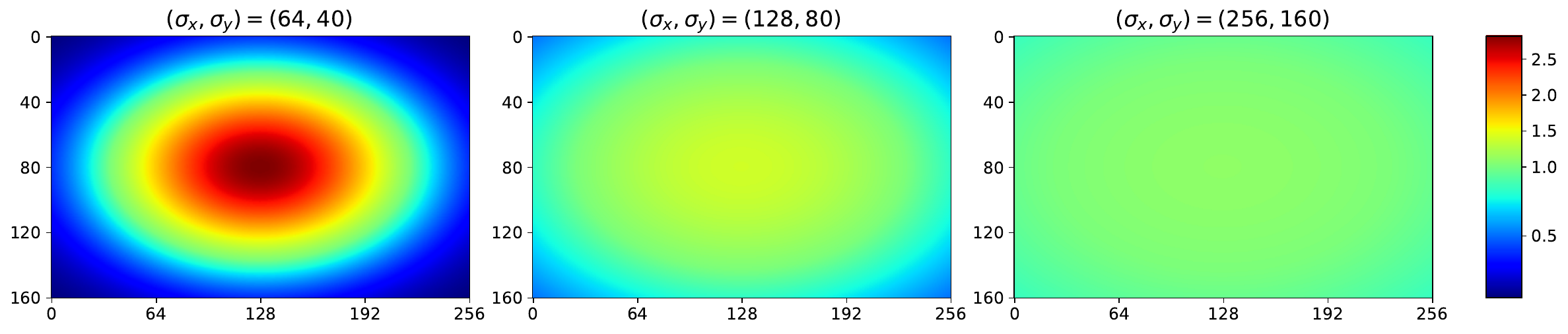}
    \vspace{-15pt}
    \caption{
    \reb{Visualization of Gaussian matrices with different standard deviations.}
     }
    \label{fig:gs_visualization}
    \vspace{-5pt}
\end{figure}

% \subsection{Sensitivity Analyses of Hyperparameters}
% \label{sec:hyperp}
We conduct sensitivity analyses on three hyperparameters:
\begin{itemize}[leftmargin=*]
\vspace{-5pt}
    \item \textbf{The long-term branch loss scale $\beta_{\text{long}}$:} As shown in \figref{fig:sensitivity_res}~({Left}), we observe that when $\beta_{\text{long}}$ for the long-term branch is too small or too large, it impedes the learning of long-term imagination, leading to a decline in performance. 
    \item \textbf{The intrinsic reward weight $\alpha$:} From \figref{fig:sensitivity_res}~({Middle}), we observe that if the hyperparameter $\alpha$ for intrinsic reward is excessively small, it may result in insufficient guidance and inaccurate reward estimation for the post-jumpy-transition state.
    \reb{\item \textbf{The intrinsic reward Gaussian parameters $(\sigma_x, \sigma_y)$:} As shown in \figref{fig:gs_visualization}, $(\sigma_x, \sigma_y)$ control the standard deviations of the Gaussian distribution along the horizontal and vertical axes, respectively. Intuitively, setting these hyperparameters too low may cause the model to overlook targets located at the edges of the observed images.
    Conversely, excessively high $(\sigma_x, \sigma_y)$ may reduce the reward discrepancy for targets at different positions within the observation, thereby diminishing the agent's incentive to focus on the target precisely. 
    From \figref{fig:sensitivity_res}~(Right), we observe that the final performance is robust to the tested parameters, with all configurations outperforming the baseline models presented in previous experiments.
    %
    % Visualizations of Gaussian matrices with varying standard deviations are provided in \figref{fig:gs_visualization}.
    % providing a clear depiction of how changes in $(\sigma_x, \sigma_y)$ values affect the distribution shapes. Thus, an optimal moderate value of $(\sigma_x, \sigma_y)$ should be selected to balance these effects.
    }
\end{itemize}

The final hyperparameters of \model{} are shown in \tabref{tab:hparams}. 

\begin{table}[ht]
\centering
\caption{Hyperparameters of \model{}.} 
\vskip 0.05in
\setlength{\tabcolsep}{8mm}{} %扩大列宽
\begin{tabular}{lccc}
\toprule
\textbf{Name} & \textbf{Notation} & \textbf{Value} \\
\midrule
\texttt{Affordance map generation}  \\

\midrule
% \reb{Dimensions of visual observation} & \reb{$W \times H$} & \reb{$256 \times 160$} \\
Sliding window size & --- & $0.15 \times 0.15$ \\
Sliding steps & --- & $9 \times 9$ \\
U-Net train epochs & --- & $500$ \\
U-Net initial learning rate & --- & $5 \times 10 ^ {-4}$ \\
U-Net learning rate decay epochs & --- & $50$ \\
U-Net learning rate decay rate & --- & $0.10$ \\
Text feature dimensions & --- & $512$ \\
\reb{Gaussian distribution standard deviations} & \reb{$\left( \sigma_x, \sigma_y \right)$} & \reb{$\left( 128, 80 \right)$} \\
\midrule
\texttt{General}  \\
\midrule
Replay capacity & --- & $1\times 10^6$ \\
Batch size & $B$ & $16$ \\
Batch length & $T$ & $32$ \\
Train ratio & --- & $16$ \\

\midrule
\texttt{World Model} \\
\midrule
Intrinsic reward weight & $\alpha$ & $1$ \\
 % $\beta_$ & --- &  \\
Deterministic latent dimensions & --- & $4{,}096$ \\
Stochastic latent dimensions & --- & $32$ \\
Discrete latent classes & --- & $32$ \\
RSSM number of units & --- & $1{,}024$ \\
World model learning rate & --- & $1 \times 10^{-4}$ \\
Long-term branch loss scale& $\beta_{\text{long}}$  & $1$ \\
Reconstruction loss scale & $\beta_{\text{pred}}$ & $1$ \\
Dynamics loss scale & $\beta_{\text{dyn}}$ & $1$ \\
Representation loss scale & $\beta_{\text{rep}}$ & $0.1$ \\

\midrule
\texttt{Behavior Learning} \\
\midrule
Imagination horizon & $L$ & $15$ \\
Discount & $\gamma$ & $0.997$ \\
$\lambda$-target  & $\lambda$ & $0.95$ \\
Actor learning rate & --- & $3\cdot10^{-5}$ \\
Critic learning rate & --- & $3\cdot10^{-5}$ \\
\bottomrule
\end{tabular}
\label{tab:hparams}
\end{table}

% \appendix
% \section{Appendix}

\end{document}